\documentclass[conference]{IEEEtran}
\usepackage{times}

\usepackage[numbers]{natbib}
\usepackage{multicol}
\usepackage[bookmarks=true]{hyperref}
\usepackage{multirow}

\usepackage{hyperref}
\usepackage{url}
\usepackage{float}

\usepackage{comment}
\usepackage{mathtools}
\usepackage{amsfonts}  %
\usepackage{cleveref}
\usepackage{graphics}
\usepackage[pdftex]{graphicx}
\usepackage{wrapfig}
\usepackage[font=footnotesize]{caption}
\usepackage{color}
\usepackage{dsfont}
\usepackage[]{mdframed}
\usepackage{algorithm}
\usepackage{algorithmic}
\usepackage{xparse}
\usepackage{amsmath}
\usepackage{bm}
\usepackage{mathtools}
\usepackage{amssymb}
\usepackage{amsthm}
\usepackage{amssymb}

\usepackage{booktabs}
\usepackage{epigraph}
\usepackage{listings}
\usepackage{xcolor} %
\usepackage{subcaption}

\lstdefinestyle{pythonstyle}{
    language=Python,
    basicstyle=\small\ttfamily,
    keywordstyle=\color{blue},
    stringstyle=\color{red},
    commentstyle=\color{green!50!black},
    morecomment=[l]{\#},
    showstringspaces=false,
    numbers=left,
    numberstyle=\tiny\color{gray},
    frame=single,
    breaklines=true,
    tabsize=4
}

\usepackage{xcolor} %
\usepackage{hyperref} %
\hypersetup{
    colorlinks=true,
}

\usepackage{xargs} %
\usepackage[colorinlistoftodos,prependcaption,textsize=tiny]{todonotes}
\newcommandx{\wrn}[2][1=]{\todo[linecolor=red,backgroundcolor=red!25,bordercolor=red,#1]{#2}}
\newcommandx{\cmt}[2][1=]{\todo[linecolor=blue,backgroundcolor=blue!25,bordercolor=blue,#1]{#2}}

\newcommand{\ba}{\mathbf{a}}

\newcommand{\bo}{\mathbf{o}}
\newcommand{\bq}{\mathbf{q}}
\newcommand{\bI}{\mathbf{I}}

\newcommand{\lang}{\ell}
\newcommand{\rawtext}{\hat{\ell}} %
\newcommand{\data}{\mathcal{D}}
\newcommand{\MM}{MM}
\newcommand{\ME}{ME}
\newcommand{\CE}{CE}
\newcommand{\WD}{WD}
\newcommand{\HL}{HL}
\newcommand{\VI}{VI}

\def \Piz {$\pi_0$}
\def \Pizfast {$\pi_0\text{-FAST}$}
\def \ModelSymbol {$\pi_{0.5}$}
\def \ModelSymbolBold {$\pi_{0.5}$}
\def \H1{H1}
\def \G1{G1}

\IEEEoverridecommandlockouts

\begin{document}
\makeatletter
\let\@oldmaketitle\@maketitle%
\renewcommand{\@maketitle}{\@oldmaketitle%
  \begin{center}
  \captionsetup{type=figure}
  \includegraphics[width=1.0\textwidth]{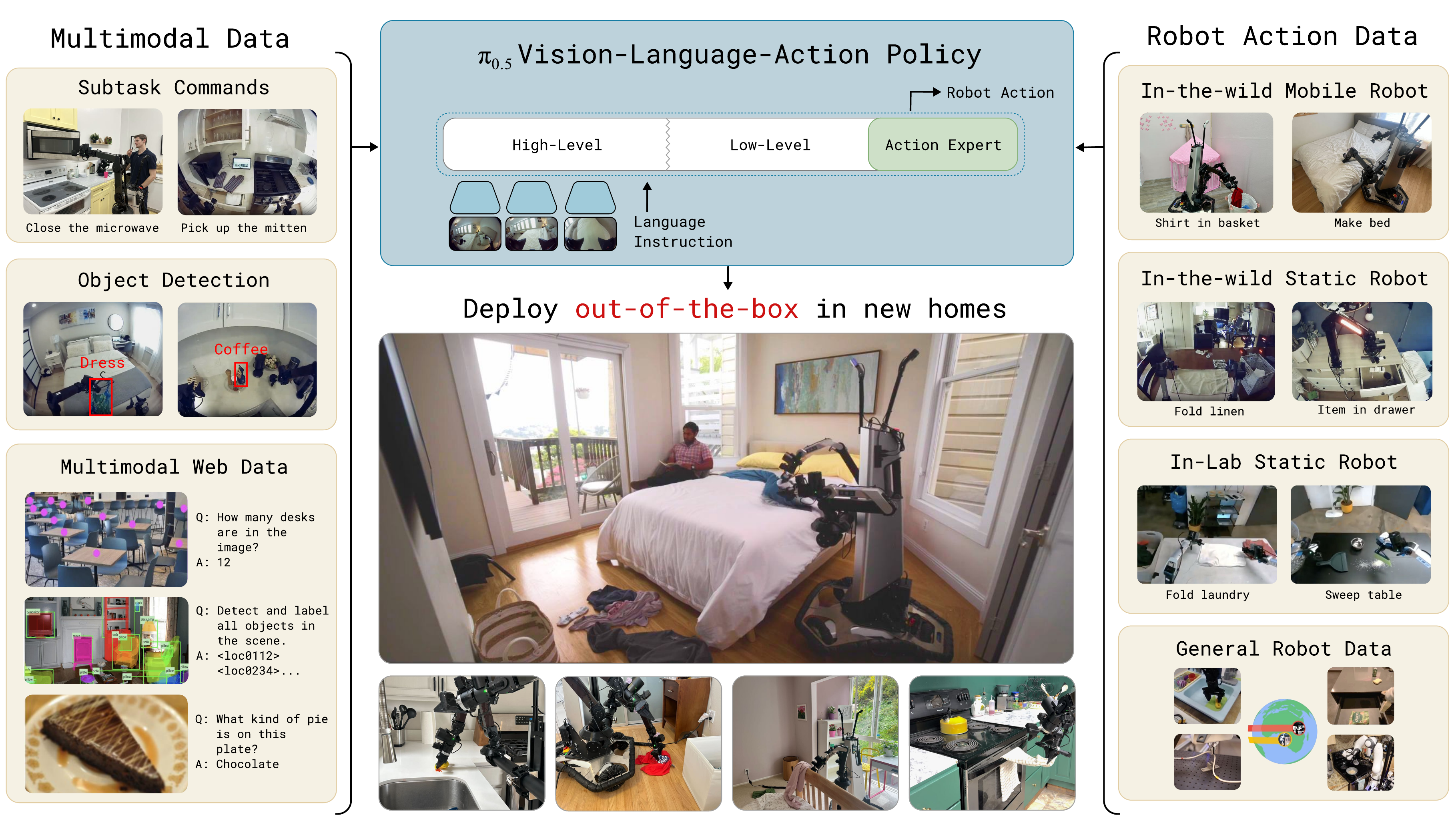}
  \caption{The \ModelSymbol\ model transfers knowledge from a heterogeneous range of data sources, including other robots, high-level subtask prediction, verbal instructions, and data from the web, in order to enable broad generalization across environments and objects. \ModelSymbol\ can control a mobile manipulator to clean kitchens and bedrooms in new homes that were not present in the training data, performing complex multi-stage behaviors with durations of 10 to 15 minutes.}
    \label{fig:teaser}
    \end{center}
}
\makeatother

\title{
\ModelSymbol: a Vision-Language-Action Model with Open-World Generalization
}

\pdfinfo{
   /Author (Physical Intelligence)
   /Title  (@title)
   /Subject (Robot Foundation Models)
   /Keywords (Robot Foundation Models)
}

\def\cameraready{0}  %

\ifx\cameraready\undefined
    \author{
    Anonymous Submission
    }
\else
    \author{
    \textbf{Physical Intelligence}\\
    \footnotesize{Kevin Black, Noah Brown, James Darpinian, Karan Dhabalia, Danny Driess, Adnan Esmail, Michael Equi,}\\
    \footnotesize{Chelsea Finn, Niccolo Fusai, Manuel Y. Galliker, Dibya Ghosh, Lachy Groom, Karol Hausman, Brian Ichter,}\\
    \footnotesize{Szymon Jakubczak, Tim Jones, Liyiming Ke, Devin LeBlanc, Sergey Levine, Adrian Li-Bell, Mohith Mothukuri,}\\
    \footnotesize{Suraj Nair, Karl Pertsch, Allen Z. Ren, Lucy Xiaoyang Shi, Laura Smith, Jost Tobias Springenberg, Kyle Stachowicz}\\
    \footnotesize{James Tanner, Quan Vuong, Homer Walke, Anna Walling, Haohuan Wang, Lili Yu, Ury Zhilinsky}\\
    \vspace{0.05in}
    \url{https://pi.website/blog/pi05}
    }
\fi

\maketitle

\begin{abstract}
In order for robots to be useful, they must perform practically relevant tasks in the real world, outside of the lab. While vision-language-action (VLA) models have demonstrated impressive results for end-to-end robot control, it remains an open question how far such models can generalize in the wild. We describe \ModelSymbol, a new model based on \Piz\ that uses co-training on heterogeneous tasks to enable broad generalization. \ModelSymbol\ uses data from multiple robots, high-level semantic prediction, web data, and other sources to enable broadly generalizable real-world robotic manipulation. Our system uses a combination of co-training and hybrid multi-modal examples that combine image observations, language commands, object detections, semantic subtask prediction, and low-level actions. Our experiments show that this kind of knowledge transfer is essential for effective generalization, and we demonstrate for the first time that an end-to-end learning-enabled robotic system can perform long-horizon and dexterous manipulation skills, such as cleaning a kitchen or bedroom, in entirely new homes.
\end{abstract}

\IEEEpeerreviewmaketitle

\section{Introduction}

\setcounter{figure}{1}

\begin{figure*}[t]
    \includegraphics[width=\linewidth]{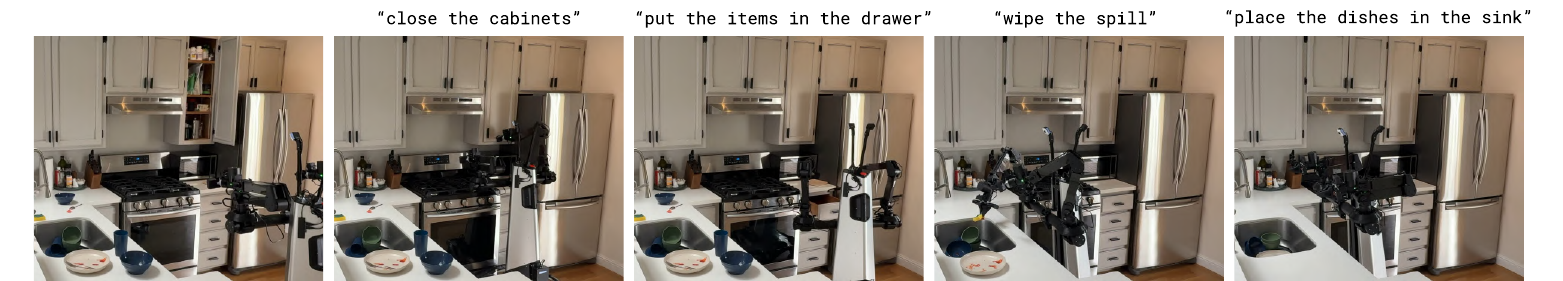}
    \caption{\textbf{\ModelSymbolBold\ cleaning a new kitchen.} The robot is tasked with cleaning a kitchen in a home that was not in the training data. The model is given general tasks (close the cabinets, put the items in the drawer, wipe the spill, and put the dishes in the sink), which it performs by both predicting subtasks to accomplish (e.g., pick up the plate) and emitting low-level actions.
    }
    \label{fig:filmstrip}
    \vspace{-0.1in}
\end{figure*}

\epigraph{\textit{Stuff your eyes with wonder... See the world. It’s more fantastic than any dream made or paid for in factories.}}{Ray Bradbury, \textit{Fahrenheit 451}}

Open-world generalization represents one of the biggest open problems in physical intelligence: embodied systems such as robotic arms, humanoids, and autonomous vehicles only truly become useful when they can leave the lab and handle the diverse situations and unexpected events that occur in the real world. Learning-based systems offer a path to enabling broad generalization, particularly with recent advances that have enabled scalable learning systems in domains ranging from natural language processing~\citep{vaswani2017attention, devlin2019bert, brown2020language, touvron2023llama} to computer vision~\citep{he2016deep, radford2021learning, he2022masked, kirillov2023segment}. However, the diversity of situations that a robot might encounter in the real world requires more than just scale: we need to design training recipes that can provide the breadth of knowledge that will allow robots to generalize at many levels of abstraction. For example, if a mobile robot is asked to clean up a kitchen that it has never seen before, some behaviors generalize readily if they are well represented in the data with a sufficient range of scenes and objects (e.g., picking up a knife or plate), others might require adapting or modifying existing skills to use them in a new way or in a new sequence, and yet others might require understanding the semantics of the scene based on prior knowledge (e.g., which drawer to open, or which object on the counter is most likely to be a drying rack). How can we structure a training recipe for a robotic learning system that can enable this kind of flexible generalization?

A person can draw on a lifetime of experience to synthesize appropriate solutions to each of these challenges. Not all of this experience is firsthand, and not all of it comes from rote practice – for example, we might use facts that we were told by others or read in a book, together with bits of insight from other tasks we have performed in different contexts, combined with direct experience in the target domain. Analogously, we might hypothesize that generalizable robotic learning systems must be able to transfer experience and knowledge from a variety of information sources. Some of these sources are firsthand experience with direct relevance to the task at hand, some require transfer from other robot embodiments, environments, or domains, and some represent entirely different data types, such as verbal instructions, perceptual tasks based on web data, or prediction of high-level semantic commands. The heterogeneity of these different sources of data present a major obstacle, but fortunately recent advances in vision-language-action (VLA) models provide us with a toolkit that can make this possible: by casting different modalities into the same sequence modeling framework, VLAs can be adapted to train on robot data, language data, computer vision tasks, and combinations of the above.

In this paper, we leverage this observation to design a co-training framework for VLAs that can utilize heterogeneous and diverse knowledge sources to enable broad generalization. Building on the \Piz\ VLA, we propose to include a range of different data sources to create the \ModelSymbol\ model (``pi oh five''), which can control mobile manipulators to perform a variety of household tasks even in homes that were never seen during training. \ModelSymbol\ draws on experience from many sources: in addition to a medium-sized dataset collected directly with mobile manipulators in a variety of real homes (about 400 hours), \ModelSymbol\ uses data from other non-mobile robots, data of related tasks collected under laboratory conditions, training examples that require predicting “high-level” semantic tasks based on robot observation, verbal language instructions provided to the robot by human supervisors, and a variety of multi-modal examples created from web data, such as image captioning, question answering, and object localization (see Figure~\ref{fig:teaser}). The overwhelming majority of training examples provided to \ModelSymbol\ (97.6\% during the first training phase) do not come from mobile manipulators performing household tasks, but from these other sources, such as other robots or data from the web. Nonetheless, \ModelSymbol\ is able to control mobile manipulators in entirely new homes not seen during training, perform intricate tasks such as hanging up towels or making beds, and can carry out long-horizon manipulation skills 10 to 15 minutes in length, cleaning an entire kitchen or bedroom based on only a high-level prompt.

The design of \ModelSymbol\ follows a simple hierarchical architecture: we first pre-train the model on the heterogeneous mixture of training tasks, and then fine-tune it specifically for mobile manipulation with both low-level action examples and high-level ``semantic'' actions, which correspond to predicting subtask labels such as ``pick up the cutting board'' or ``rearrange the pillow.'' At runtime, during each step of inference, the model first predicts the semantic subtask, inferring the behavior that is appropriate to perform next based on the task structure and the semantics of the scene, and then predicts the low-level robot action chunk based on this subtask. This simple architecture provides both the ability to reason about long-horizon multi-stage tasks and the ability to leverage different sources of knowledge for the two levels: the low-level action inference procedure readily benefits from action data collected by other robots, including simpler static robots in other environments, while the high-level inference procedure benefits from semantic examples from the web, high-level annotation prediction, and even verbal commands that can be provided to the robot by human ``supervisors'' that walk the robot through complex tasks step by step, instructing it (much like how they might instruct a person) on the appropriate subtasks to perform to complete a complex task such as cleaning a room. We illustrate this design in Figure~\ref{fig:teaser}.

Our central contribution is a system for training a highly generalizable VLA, \ModelSymbol, together with a proof of concept that generalization can emerge from this model when it is trained on appropriately diverse data. We provide a detailed empirical evaluation of both \ModelSymbol's generalization capabilities and the relevance of different co-training ingredients. To our knowledge, our work is the first to demonstrate an end-to-end learning-enabled robotic system that can perform long-horizon and dexterous manipulation skills, such as cleaning a kitchen or bedroom, in entirely new homes. Our experiments and comparisons further show that this is enabled by transferring knowledge from other robots, high-level semantic prediction, verbal language instruction from human supervisors, web data, and other sources.

\section{Related Work}
\label{sec:related}

\noindent\textbf{Generalist robot manipulation policies.}
Recent works have demonstrated that broadening the training data distribution for robot manipulation policies from narrow, single-task datasets to diverse datasets that span many scenes and tasks~\citep{dasari2019robonet, ebert2021bridge, walke2023bridgedata, open_x_embodiment_rt_x_2023, khazatsky2024droid, bharadhwaj2023roboagent, fang2024rh20t, shafiullah2023bringingrobotshome, contributors2025agibotworld} allows the resulting policies to not only solve a wider range of tasks out of the box, but also improves their ability to generalize to \emph{new} scenes and tasks~\citep{rt12022arxiv, open_x_embodiment_rt_x_2023, octo_2023, Doshi24-crossformer}. Training such \emph{generalist} policies requires new modeling approaches that can handle the scale and diversity of datasets that often span hundreds of different tasks and scenes. Vision-language-action models (VLAs)~\citep{driess2023palm, rt22023arxiv, kim2024openvla, black2024pi_0, wen2024tinyvlafastdataefficientvisionlanguageaction, zhen20243dvla, liu2024rdt, li2024cogact, belkhale2024minivla, szot2024multimodal, pertsch2025fast, geminirobotics2025, wen2025dexvla, bjorck2025gr00t, huang2025otter} offer an appealing solution: by fine-tuning pre-trained vision-language models for robot control, VLAs can leverage the semantic knowledge acquired from web-scale pretraining and bring it to bear on the robotics problem. When combined with highly expressive action decoding mechanisms like flow matching~\citep{black2024pi_0}, diffusion~\citep{liu2024rdt, wen2025dexvla, liu2025hybridvla}, or advanced action tokenization schemes~\citep{pertsch2025fast}, VLAs can perform a wide range of complex manipulation tasks in the real world. However, despite impressive language following abilities, VLAs are still typically evaluated in environments that closely match their training data. While some studies suggest that simple skills like picking up objects or opening drawers can be made to generalize simply by collecting robot data in a broader set of environments~\citep{chi2024universal, shafiullah2023bringingrobotshome, etukuru2024robotutilitymodelsgeneral, lin2024data, pertsch2025fast}, it is challenging to apply the same approach to more complex, long-horizon tasks like cleaning up a kitchen, where achieving broad coverage of plausible scenarios via brute-force scaling of robot data collection is infeasible. In our experiments, we evaluate \ModelSymbol\ in entirely new scenes, such as new kitchens and bedrooms that were not seen in training, showing that our VLA can generalize to entirely new scenes by leveraging not only direct first-hand experience on the target mobile manipulator platform, but also information from other data sources. These sources include data from other (non-mobile) robots, high-level semantic subtask prediction, and data from the web.

\begin{figure*}[t]
    \centering
    \includegraphics[width=1.0\linewidth]{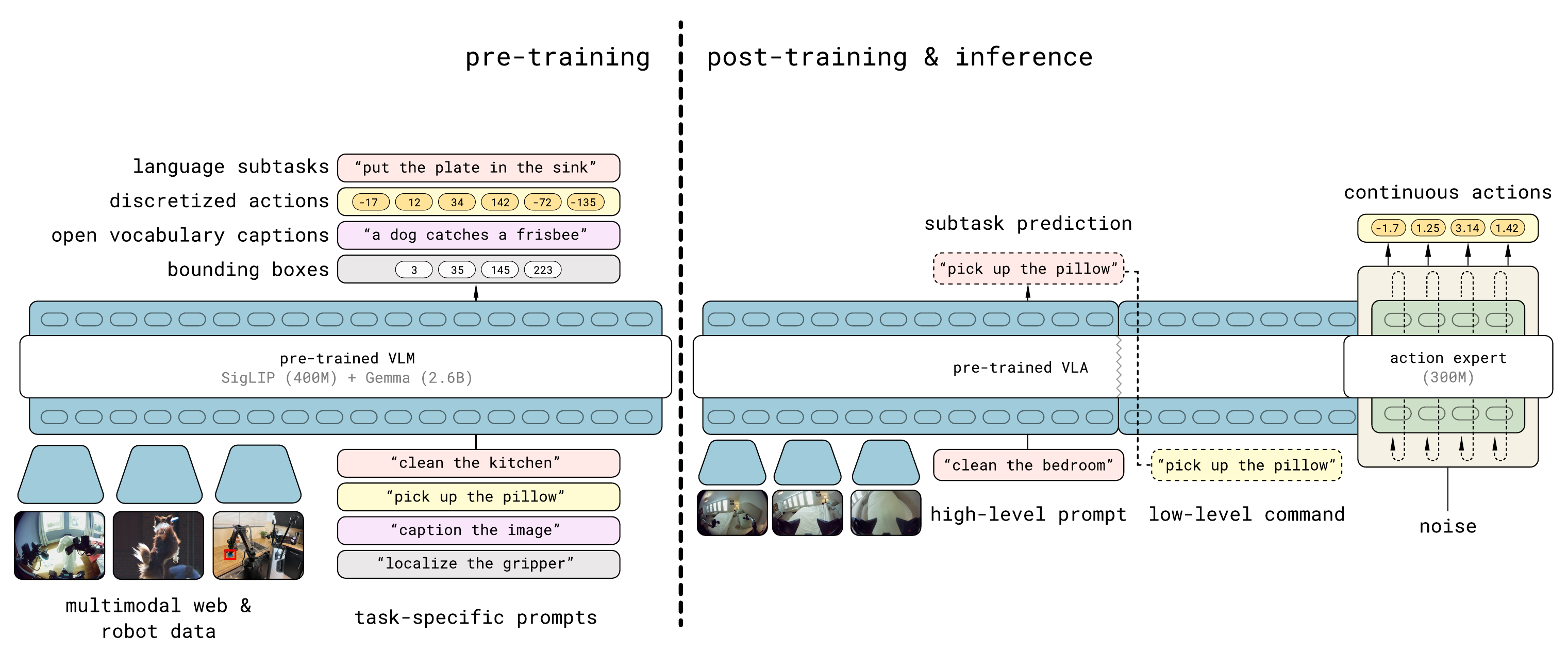}
    \caption{\textbf{Model overview.} \ModelSymbol\ is trained in two stages. First, a pre-training stage combines all of the different data sources to produce an initial VLA with discrete tokens. This stage uses data from diverse robotic platforms, high-level semantic action prediction, and data from the web. Robotic data uses the FAST action tokenizer to represent actions as discrete tokens~\citep{pertsch2025fast}. Second, a post-training stage specializes the model for low-level and high-level inferences for mobile manipulation, leveraging the most task-relevant data, including verbal instructions from human supervisors. This stage uses flow matching to represent the action distribution, enabling efficient real-time inference and the ability to represent fine-grained continuous action sequences. At inference time, the model first infers a high-level subtask, and then predicts the actions based on this subtask.}
    \label{fig:overview}
\end{figure*}

\noindent\textbf{Non-robot data co-training.}
A number of prior works have sought to use diverse \emph{non-robot} data to improve the generalization of robot policies. Prior methods have explored initializing vision encoders from computer vision datasets~\citep{xiao2022masked, nair2022r3m, majumdar2023we, dasari2023unbiased}, or leveraging off-the-shelf task planners~\citep{huang2022language, liang2023code, singh2023progprompt, wang2024llm}. VLA policies are typically initialized from a pre-trained vision-language model, which has been exposed to large amounts of internet vision and language data~\citep{driess2023palm, rt22023arxiv, kim2024openvla}. Notably, the VLA architecture is flexible and allows to map between input and output sequences of multi-modal vision, language, and action tokens. As such, VLAs broaden the design space of possible transfer approaches beyond simple weight initialization, by supporting the \emph{co-training} of a single, unified architecture on not just robot action imitation data, but any dataset that interleaves one or multiple of the aforementioned modalities. Prior works have demonstrated that co-training VLAs with data mixtures used for VLM training~\citep{driess2023palm, rt22023arxiv, yang2025magma} can improve their generalization ability, e.g., when interacting with new objects or unseen scene backgrounds. In this work, we go beyond VLM data co-training and design a system for co-training VLAs with a broader set of robotics-relevant supervision sources, including data from other robots, high-level semantic subtask predictions, and verbal language instructions. While multitask training and co-training are not new ideas, we show that the specific combination of data sources in our system enables mobile robots to perform complex and long-horizon behaviors in entirely new environments. We believe that this level of generalization, particularly when accounting for the complexity of the tasks, goes significantly beyond the results demonstrated in prior works.

\noindent \textbf{Robot reasoning and planning with language.}
A number of prior works have shown that augmenting end-to-end policies with high-level reasoning can significantly improve performance for long-horizon tasks~\citep{saycan2022arxiv, hu2023look, li2023itp, stone2023open, shi2024yell,belkhale2024rthactionhierarchiesusing, dai2024racer, chen2024automating, liu2024ok, Zawalski24-ecot, liu2024moka, nasiriany2024pivot, cheng2024navila, shah2024bumble, zhi2024closed, qiu2024open, shi2025hi, li2025hamster, geminirobotics2025, zhao2025cot}, particularly when high-level subtask inference can benefit from large pre-trained LLMs and VLMs. Our method also uses a two-stage inference procedure, where we first infer a high-level semantic subtask (e.g., ``pick up the plate''), and then predict the action based on this subtask. Many prior methods have employed two separate models for this purpose, with a VLM predicting semantic steps and a separate low-level policy executing those steps~\citep{saycan2022arxiv, shi2024yell, cheng2024navila, duan2024manipulate, shah2024bumble, shi2025hi, li2025hamster}. Our method uses the same exact model for both high-level and low-level inference, in a recipe that more closely resembles chain-of-thought~\citep{wei2022chain} or test-time compute~\citep{jaech2024openai} methods, though unlike embodied chain-of-thought methods~\citep{Zawalski24-ecot, li2024llara, niu2024llarva}, the high-level inference process still runs at a lower frequency than low-level action inference.

\noindent \textbf{Robotic learning systems with open-world generalization.} While most robotic learning systems are evaluated in environments that closely match the training data, a number of prior works have explored broader open-world generalization. When the robot's tasks are restricted to a more narrow set of basic primitives, such as picking up objects, methods that allow for task-specific assumptions (e.g., grasp prediction, or incorporating model-based planning and control) have been shown to generalize broadly, even to entirely new homes~\citep{jones2006robots, dempsey2023reviews, nguyen2014autonomously, mahler2017dex, fang2023anygrasp}. However, such methods do not readily generalize to the full range of possible tasks that a generalist robot might need to perform. More recently, large-scale datasets collected across many domains~\citep{khazatsky2024droid, shahGNMGeneralNavigation2023, open_x_embodiment_rt_x_2023, shafiullah2023bringingrobotshome, chi2024universal, lin2024data} have been shown to enable generalization of simple but end-to-end learned tasks to new environments~\citep{gupta2018robot, gervet2023navigating, shafiullah2023bringingrobotshome, shah2023vint, ehsani2023spoc, lin2024data, etukuru2024robotutilitymodelsgeneral, pertsch2025fast}. However, the tasks in these demonstrations are still relatively simple, typically less than a minute in length and often with relatively low success rates. We show that \ModelSymbol\ can perform long, multi-stage tasks, such as putting all of the dishes in the sink or picking all of the clothing off the floor of a new bedroom, while generalizing to entirely new homes.

\section{Preliminaries}
\label{sec:preliminaries}

Vision-language-action models (VLAs) are typically trained via imitation learning on diverse robot demonstration datasets~$\data$, by maximizing the log-likelihood of an action $\ba_t$ (or, more generally, an action \emph{chunk} $\ba_{t:t+H}$) given an observation~$\bo_t$ and a natural language task instruction~$\lang$: $\max_\theta \mathbb{E}_{(\ba_{t:t+H}, \bo_t, \lang) \sim \mathcal{D}} \log \big(\pi_\theta(\ba_{t:t+H} \vert \bo_t, \lang)\big)$. The observation typically contains one or more images $\bI^1_t, ..., \bI^n_t$ and proprioceptive state $\bq_t$, which captures the position of the robot's joints.
VLA architectures follow the design of modern language and vision-language models, with modality-specific tokenizers that map inputs and outputs to discrete (``hard'') or continuous (``soft'') token representations, and a large, auto-regressive transformer backbone that is trained to map from input to output tokens. The weights of these models are initialized from pre-trained vision-language models. By encoding policy inputs and outputs into tokenized representations, the imitation learning problem described above can be cast as a simple next-token-prediction problem over a sequence of observation, instruction and action tokens, and we can leverage the scalable tools of modern machine learning to optimize it. In practice, the choice of tokenizers for image and text inputs follows those of modern vision-language models. For actions, prior work has developed effective, compression-based tokenization approaches~\citep{pertsch2025fast}, which we use in this work during pretraining. A number of recent VLA models have also proposed to represent the action distribution via diffusion~\citep{liu2024rdt, wen2025dexvla, liu2025hybridvla} or flow matching~\citep{black2024pi_0}, providing a more expressive representation over continuous-valued action chunks. During the post-training phase of our model, we will build on the design of the \Piz\ model~\citep{black2024pi_0}, which represents the action distribution via flow matching. In this design, the tokens corresponding to actions receive the partially denoised actions from the previous step of flow matching as input, and output the flow matching vector field. These tokens also use a different set of model weights, which we refer to as an ``action expert,'' analogously to a mixture of experts architecture. This action expert can specialize to flow matching-based action generation, and can be significantly smaller than the rest of the LLM backbone.

\section{The \ModelSymbol\ Model and Training Recipe}
\label{sec:model}

We provide an overview of the \ModelSymbol\ model and training recipe in Figure~\ref{fig:overview}.
The model weights are initialized from a standard VLM trained on data from the web, and training then proceeds in two stages: a pre-training stage intended to adapt the model to diverse robotic tasks, and a post-training stage intended to specialize it to mobile manipulation and equip it with the mechanisms for efficient test-time inference. During pre-training, all tasks, including tasks with robot actions, are represented with discrete tokens, which leads to simple, scalable, and efficient training~\citep{pertsch2025fast}. During post-training, we adapt the model to also have an action expert, as with \Piz, in order to both represent actions with finer granularity and enable more compute-efficient inference for real-time control. At inference-time, the model first produces a high-level subtask for the robot to perform and then, conditioned on this subtask, predicts the low-level actions via the action expert. We describe the model architecture below, followed by a description of each of the phases and their corresponding training tasks.

\subsection{The \ModelSymbol\ architecture}

The \ModelSymbol\ architecture can flexibly represent both action chunk distributions and tokenized text outputs, with the latter used both for co-training tasks (e.g., question-answering) and for outputting high-level subtask predictions during hierarchical inference. The distribution captured by the model can be written as $\pi_\theta(\ba_{t:t+H}, \rawtext \vert \bo_t, \lang)$, where $\bo_t = [\bI^1_t, ..., \bI^n_t, \bq_t]$ consists of the images from all of the cameras and the robot's configuration (joint angles, gripper pose, torso lift pose, and base velocity), $\lang$ is the overall task prompt (e.g., ``put away the dishes''), $\rawtext$ represents the model's (tokenized) textual output, which could be either a predicted high-level subtask (e.g., ``pick up the plate'') or the answer to a vision-language prompt in web data, and $\ba_{t:t+H}$ is a predicted action chunk. We decompose the distribution as
\begin{equation*}
\pi_\theta(\ba_{t:t+H}, \rawtext \vert \bo_t, \lang) = \pi_\theta(\ba_{t:t+H} \vert \bo_t, \rawtext)\pi_\theta(\rawtext \vert \bo_t, \lang),
\end{equation*}
where the action distribution does not depend on $\lang$, only on $\rawtext$. Thus, high-level inference captures $\pi_\theta(\rawtext \vert \bo_t, \lang)$, and low-level inference captures $\pi_\theta(\ba_{t:t+H} \vert \bo_t, \rawtext)$, with both distributions represented by the same model.

The model corresponds to a transformer that takes in $N$ multimodal input tokens $x_{1:N}$ (we use the term token loosely here, referring to both discretized and continuous inputs) and produces a sequence of multimodal outputs $y_{1:N}$, which we can write as $y_{1:N} = f\big(x_{1:N}, A(x_{1:N}), \rho(x_{1:N})\big)$. Each $x_i$ can be a text token ($x_i^w\in\mathbb{N}$), an image patch ($x_i^I\in\mathbb{R}^{p\times p \times 3}$), or an intermediate denoising value of a robot action in flow matching ($x_i^a\in\mathbb{R}^d$).
The observations $\bo_t$ and $\lang$ form the prefix part of $x_{1:N}$.
Depending on the token type, as indicated by $\rho(x_i)$, each token can be processed not only by a different encoder, but also by different expert weights within the transformer.
For example, image patches are fed through a vision encoder, and text tokens are embedded with an embedding matrix.
Following \Piz~\citep{black2024pi_0}, we linearly project action tokens $x_i^a$ into the transformer embedding space and use separate expert weights in the transformer to process the action tokens.
The attention matrix $A(x_{1:N})\in[0,1]^{N\times N}$ indicates if a token can attend to another token.
Compared to standard causal attention in LLMs, image patch, textual prompt, and continuous action tokens use bidirectional attention.

As we want our model to output both text (to answer questions about the scene or to output next tasks to accomplish) and actions (to act in the world), the output of $f$ is split into text token logits and action output tokens, respectively $\big(y^\ell_{1:M}, y^a_{1:H}\big)$.
The first $M$ correspond to text token logits that can be used to sample $\rawtext$ and the later $H$ tokens are produced by a separate action expert, as in \Piz, and projected via a linear mapping to continuous outputs used to obtain $\ba_{t:t+H}$ (see next section).
Note that $M + H \le N$, i.e., not all outputs are associated with a loss.
The robot proprioceptive state is discretized and input to the model as text tokens.
More details about the architecture are in Appendix~\ref{app:model}.

\subsection{Combining discrete \& continuous action representations}

\begin{figure*}
    \centering
    \includegraphics[width=\linewidth]{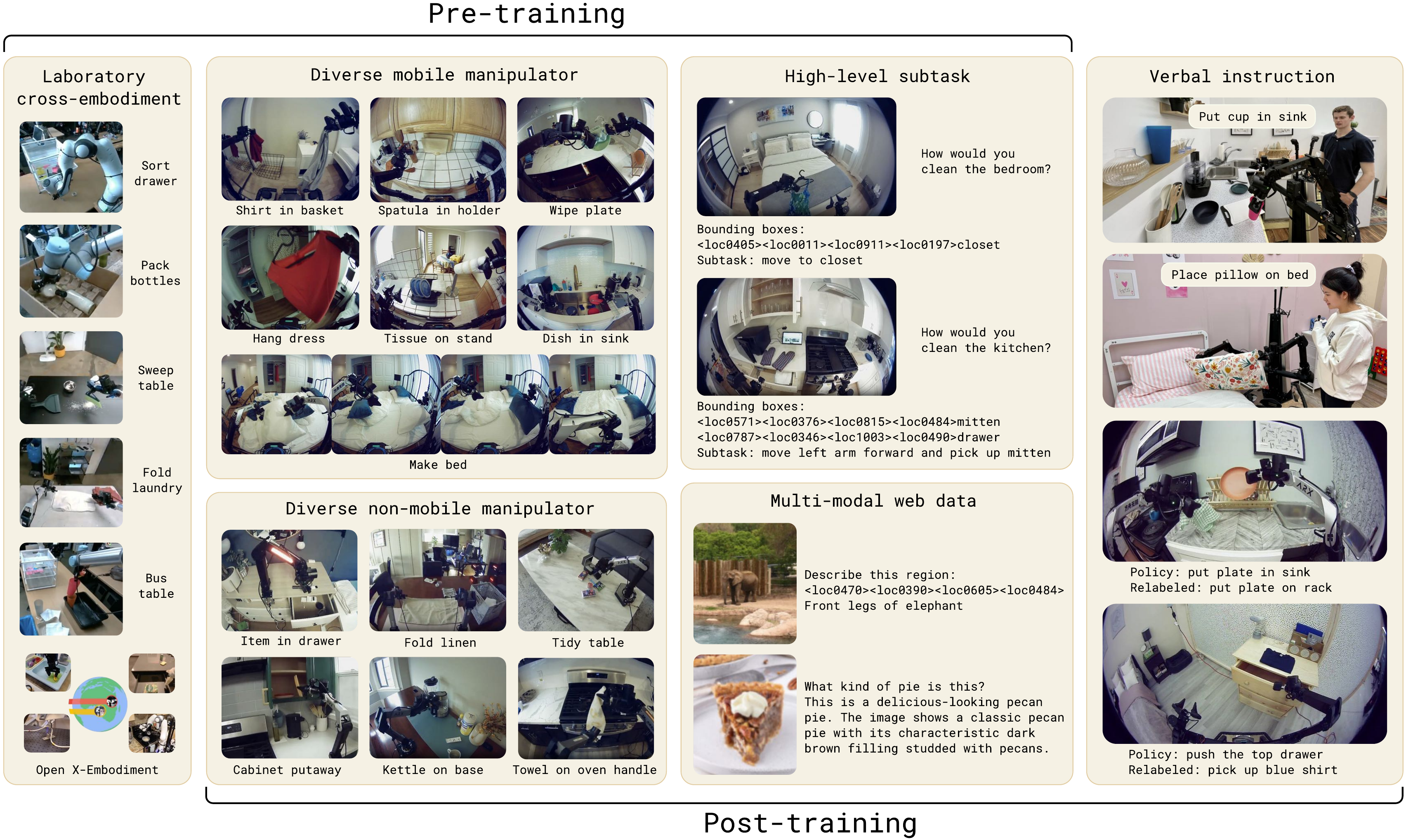}
    \caption{\textbf{Examples from pre-training and post-training tasks.} \ModelSymbol\ is pre-trained on data from mobile manipulators (\texttt{MM}), non-mobile robots in diverse environments (\texttt{ME}), and cross-embodiment data collected under laboratory conditions (\texttt{CE}), as well as high-level subtask prediction (\texttt{HL}), and multi-modal web data (\texttt{WD}). In a post-training phase, we additionally use verbal instructions (\texttt{VI}), and omit the laboratory cross-embodiment data (\texttt{CE}) to focus the model on mobile manipulation and diverse environments. The figure displays an exemplary subset of the tasks in each category.
    }
    \label{fig:pretraining}
\end{figure*}

Similarly to \Piz, we use flow-matching~\citep{lipman2022flow} to predict continuous actions in the final model. Given $\ba_{t:t+H}^{\tau, \omega} = \tau \ba_{t:t+H} + (1-\tau)\omega$, $\omega \sim \mathcal{N}(0,\bI)$, where $\tau\in[0,1]$ is the flow matching time index, the model is trained to predict the flow vector field $\omega - \ba_{t}$. However, as shown in \citep{pertsch2025fast}, VLA training can be much faster when actions are represented by discrete tokens, particularly when using a tokenization scheme that is efficient for compressing the action chunks (e.g., FAST). Unfortunately, such discrete representations are less well-suited for real-time inference, because they require expensive autoregressive decoding for inference~\citep{pertsch2025fast}. Therefore, an ideal model design would train on discretized actions but still allow for use of flow matching to produce continuous actions at inference time.

Our model is therefore trained to predict actions \emph{both} through autoregressive sampling of tokens (using the FAST tokenizer) and iterative integration of the flow field, combining the best of both worlds.
We use the attention matrix to ensure that the different action representations do not attend to each other.
Our model is optimized to minimize the combined loss
\begin{align}
    \mathbb{E}_{\mathcal{D}, \tau, \omega} \Big[&\ \!H\big(x_{1:M}, f^\lang_\theta(\bo_t, \lang)\big) \notag \\
    & + \alpha \left\|\omega - \ba_{t:t+H} - f^a_\theta(\ba^{\tau, \omega}_{t:t+H}, \bo_t, \lang)\right\|^2 \Big], \label{eq:cotrain}
\end{align}
where $H(x_{1:M}, y^\lang_{1:M})$ is the cross entropy loss between the text tokens and predicted logits (including the FAST encoded action tokens), $y^a_{1:H} = f^a_\theta(\ba^{\tau, \omega}_{t:t+H}, \bo_t, \lang)$ is the output from the (smaller) action expert, and $\alpha\in\mathbb{R}$ is a trade-off parameter.
This scheme enables us to first pre-train our model as a standard VLM transformer model by mapping actions to text tokens ($\alpha=0$), and then add additional action expert weights predicting continuous action tokens in a non-autoregressive fashion for fast inference in a post-training stage.
We find that following this procedure, which is further explained below, leads to stable pre-training and excellent language following abilities of the VLA model. At inference time we then use standard autoregressive decoding for text tokens $\hat{\lang}$ followed by $10$ denoising steps, conditioned on text tokens, to produce actions $\ba_{t:t+H}$.

\subsection{Pre-training}

In the first training stage, \ModelSymbol\ is trained with a broad range of robot and non-robot data, which we summarize below and illustrate in Figure~\ref{fig:pretraining}. It is trained as a standard auto-regressive transformer, performing next-token prediction of text, object locations, and FAST encoded action tokens. 

\noindent \textbf{Diverse \emph{M}obile \emph{M}anipulator data (\MM).} We use about 400 hours of data of mobile manipulators performing household tasks in about 100 different home environments, some of which are shown in Figure~\ref{fig:realhomes}, using the robots in Section~\ref{sec:datadetails}. This slice of the training set is the most directly relevant to our evaluation tasks, which consist of similar cleaning and tidying tasks in new, unseen, home environments.

\noindent \textbf{Diverse \emph{M}ulti-\emph{E}nvironment non-mobile robot data (\ME).} We also collected non-mobile robot data, either with a single arm or two arms, in a variety of home environments. These arms were fixed to surfaces or mounting platforms, and because they are significantly lighter and easier to transport, we were able to gather a more diverse dataset in a wider range of homes with them. However, this \ME\ data comes from a different embodiment than the mobile robots.

\noindent \textbf{\emph{C}ross-\emph{E}mbodiment laboratory data (\CE).} We collected data for a wide range of tasks (e.g., bussing a table, folding shirts) in the laboratory, with simpler tabletop environments and a variety of robot types. Some of these tasks are highly relevant to our evaluation (e.g., putting dishes in a bin), while others are not (e.g., grinding coffee beans). This data includes single-arm and dual-arm manipulators, and both static and mobile bases. We also include the open-source OXE dataset \cite{collaboration2023open}. This dataset is an extended version of the dataset used by \Piz \citep{black2024pi_0}.

\noindent \textbf{\emph{H}igh-\emph{L}evel subtask prediction (\HL).} Breaking down high-level task commands such as ``clean the bedroom'' into shorter subtasks like ``adjust the blanket'' and ``pick up pillow'', similar to chain-of-thought prompting for language models, can help a trained policy reason about the current scene and better determine the next action. For robot data in \MM, \ME, and \CE\ where the task involves multiple subtasks, we manually annotate all data with semantic descriptions of the subtasks and train \ModelSymbol\ to jointly predict the subtask labels (as text) as well as the actions (conditioned on the subtask label) based on the current observation and high-level command. This naturally leads to a model that can act both as a high-level policy (outputting subtasks) and low-level policy that executes actions for these subtasks. We also label relevant bounding boxes shown in the current observation and train \ModelSymbol\ to predict them before predicting the subtask.

\noindent \textbf{Multi-modal \emph{W}eb \emph{D}ata (\WD).} Finally we include a diverse set of web data involving image captioning (CapsFusion \cite{yu2024capsfusion}, COCO \cite{chen2015microsoft}), 
question answering (Cambrian-7M \cite{tong2024cambrian}, PixMo \cite{deitke2024molmo}, VQAv2 \cite{goyal2017making}), and object localization in pre-training. For object localization, we further extend the standard datasets with additional web data of indoor scenes and household objects with bounding box annotations.

For all action data, we train the model to predict target joint and end-effector poses. To differentiate the two, we add `$<$control\_mode$>$ joint/end effector $<$control\_mode$>$' to the text prompt. 
All action data is normalized to $[-1,1]$ using the 1\% and 99\% quantile of each action dimension of the individual dataset. 
We set the dimensionality of the action $\ba$ to a fixed number to accommodate the largest action space among all the datasets.
For robots with lower-dimensional configuration and action spaces, we zero-pad the action vectors.

\subsection{Post-training}

After pre-training the model with discrete tokens for 280k gradient steps, we perform a second stage of training that we refer to as post-training.
The purpose of this stage is to both specialize the model to our use-case (mobile manipulation in homes), and to add an action expert that can produce continuous action chunks via flow matching. This stage jointly trains with next-token prediction, to preserve text prediction capabilities, and flow matching for the action expert (which is initialized with random weights at the beginning of post-training).
We optimize the objective in Equation~\eqref{eq:cotrain}, with $\alpha = 10.0$ for 80k additional steps. The post-training action dataset consists of the \MM\ and \ME\ robot data, filtered down to successful episodes that are below a fixed length threshold. We include web data (\WD) to preserve the model's semantic and visual capabilities, and the slice of \HL\ data corresponding to the multi-environment datasets. Additionally, to improve the model's ability to predict appropriate high-level subtasks, we collect \emph{verbal instruction} demonstrations (\VI), which are constructed by expert users providing ``language demonstrations,'' selecting appropriate sub-task commands to command the robot to perform mobile manipulation tasks step by step. These examples are collected by ``teleoperating'' the robot in real time with language to perform tasks with the learned low level policy, essentially providing demonstrations of good high-level subtask outputs for a trained policy.

\begin{figure}
    \centering
    \includegraphics[width=\linewidth]{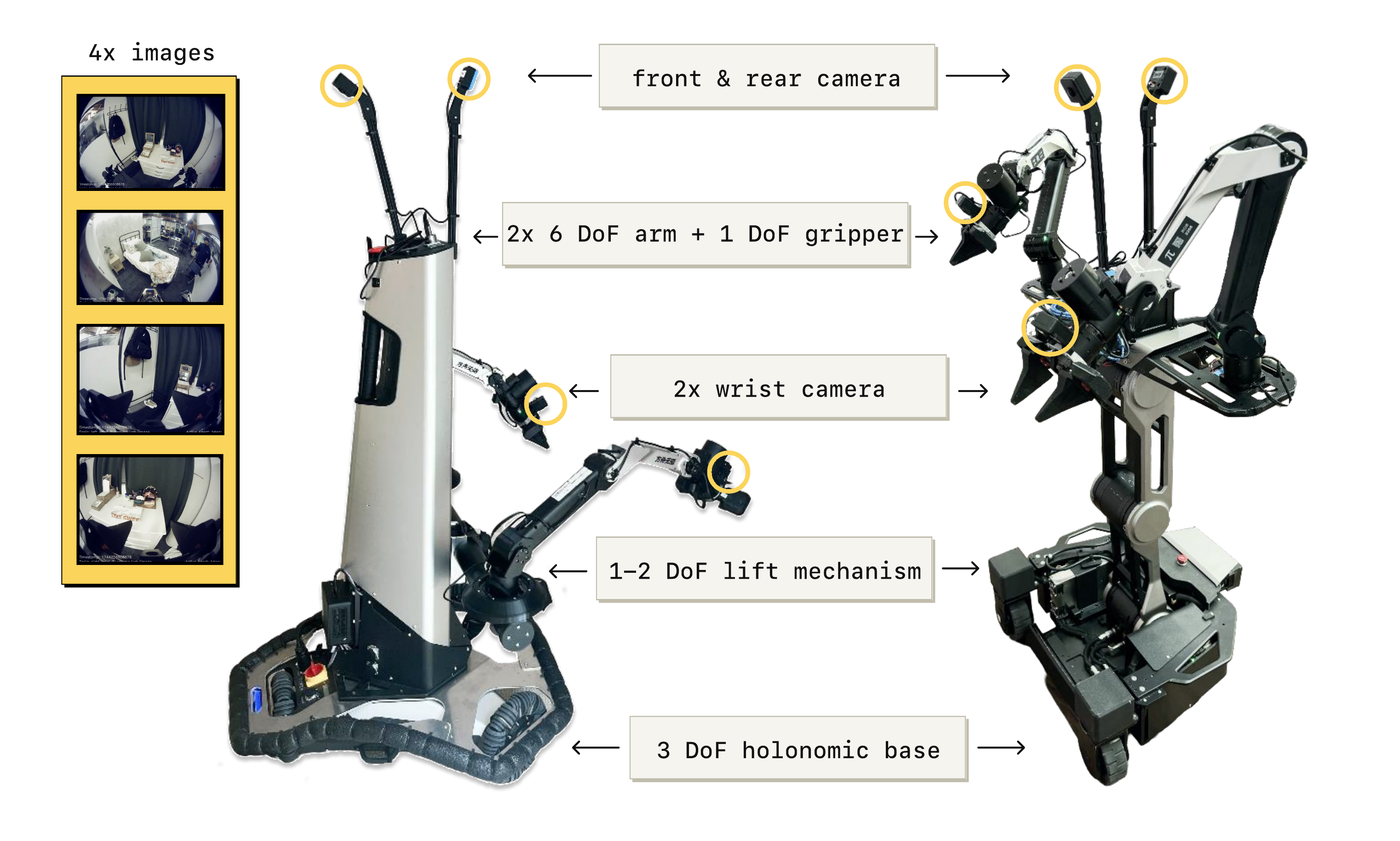}
    \caption{\textbf{Robot system overview.} We use two mobile manipulator platforms -- each has four cameras (forward, backward, and both wrists), two 6 DoF arms with parallel jaw grippers, a mobile base, and a torso lift mechanism. The \ModelSymbol\ model controls the joints and grippers of each arm, base velocity, and the lift position, resulting in 18-19 DoF state and action spaces.}
    \label{fig:robot}
\end{figure}

\subsection{Robot system details}
\label{sec:datadetails}

The robot systems used in our mobile manipulation experiments are illustrated in Figure~\ref{fig:robot}. We conducted all of our experiments using two types of mobile manipulators. Both platforms are equipped with two 6 DoF arms with parallel jaw grippers and wrist-mounted monocular RGB cameras, a wheeled holonomic base, and a torso lift mechanism. The state and action spaces for the base correspond to linear (2D) and angular (1D) velocity, and the torso lift mechanism is either 1D (up/down) or 2D (up/down and forward/backward). In addition to the two wrist cameras, the robots have a forward and backward facing camera mounted between the arms. We use all four cameras for high-level inference, and the wrist and forward cameras for the low-level inference process. The total dimensionality of the state and action spaces is 18 or 19, depending on the platform.

The control system is very simple: the \ModelSymbol\ model directly commands target poses for the arms, gripper, and torso lift, and the target base velocities at 50 Hz (with action chunking). These targets are tracked with simple PD controllers, without any additional trajectory planning or collision detection. All manipulation and navigation control is fully end-to-end.

\section{Experimental Evaluation}
\label{sec:experiments}

\begin{figure*}
\centering
\includegraphics[width=.95\linewidth]{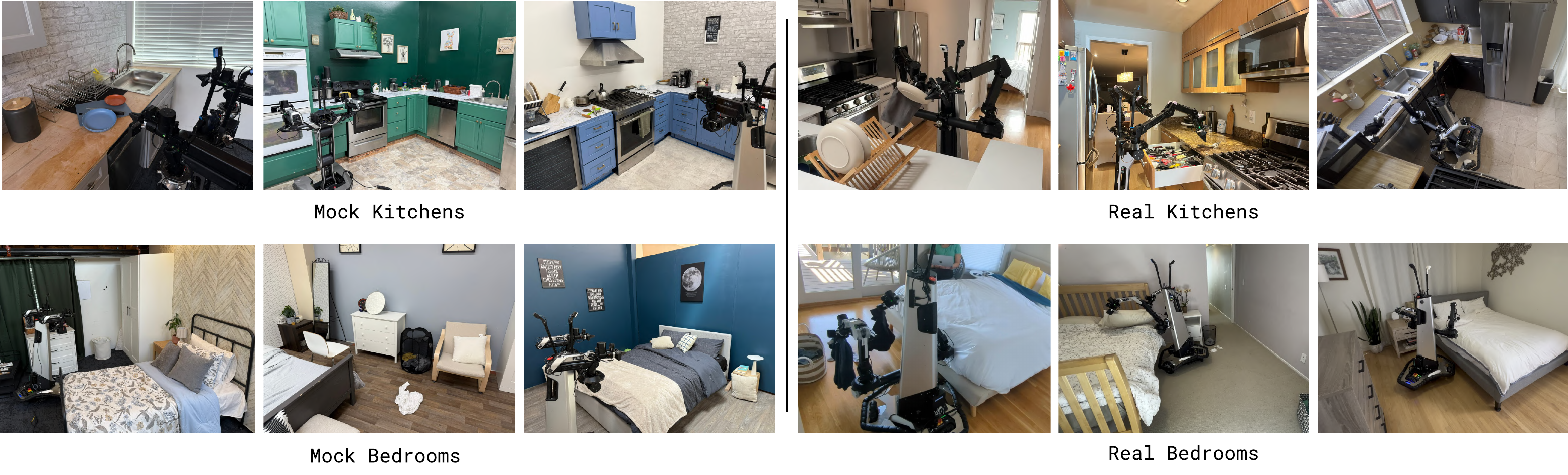}
    \caption{\textbf{Evaluation environments.} We evaluate \ModelSymbol\ in entirely new kitchens and bedrooms that were not seen during training, with novel objects, backgrounds, and layouts. We use a set of mock rooms for controlled, reproducible quantitative comparisons (left) and real homes for a realistic final evaluation (right).}
    \label{fig:mock_envs}
\end{figure*}

The \ModelSymbol\ model is designed to generalize broadly to new environments. While it is common to evaluate VLAs in environments that match the training data, we conduct all of our experiments in novel environments that were not seen in training. For quantitative comparisons, we use a set of mock home environments to provide a controlled and reproducible setup, while the most realistic final evaluation is conducted in three real homes that were not part of the training set (see Figure~\ref{fig:mock_envs}). Our experiments focus on the following questions:

\begin{enumerate}
    \item Can \ModelSymbol\ effectively generalize to complex multi-stage tasks in entirely new homes?
    \item How does the generalization of \ModelSymbol\ scale with the number of distinct environments in the training data?
    \item How do the individual co-training ingredients in the \ModelSymbol\ training mixture contribute to its final performance?
    \item How does \ModelSymbol\ compare to the \Piz\ VLA?
    \item How important is the high-level inference component of \ModelSymbol, and how does it compare to flat, low-level inference as well as oracle high-level baselines?
\end{enumerate}

\subsection{Can \ModelSymbol\ generalize to real homes?}

\begin{figure*}
    \centering
    \begin{subfigure}{0.57\textwidth}
        \centering
        \includegraphics[width=\linewidth]{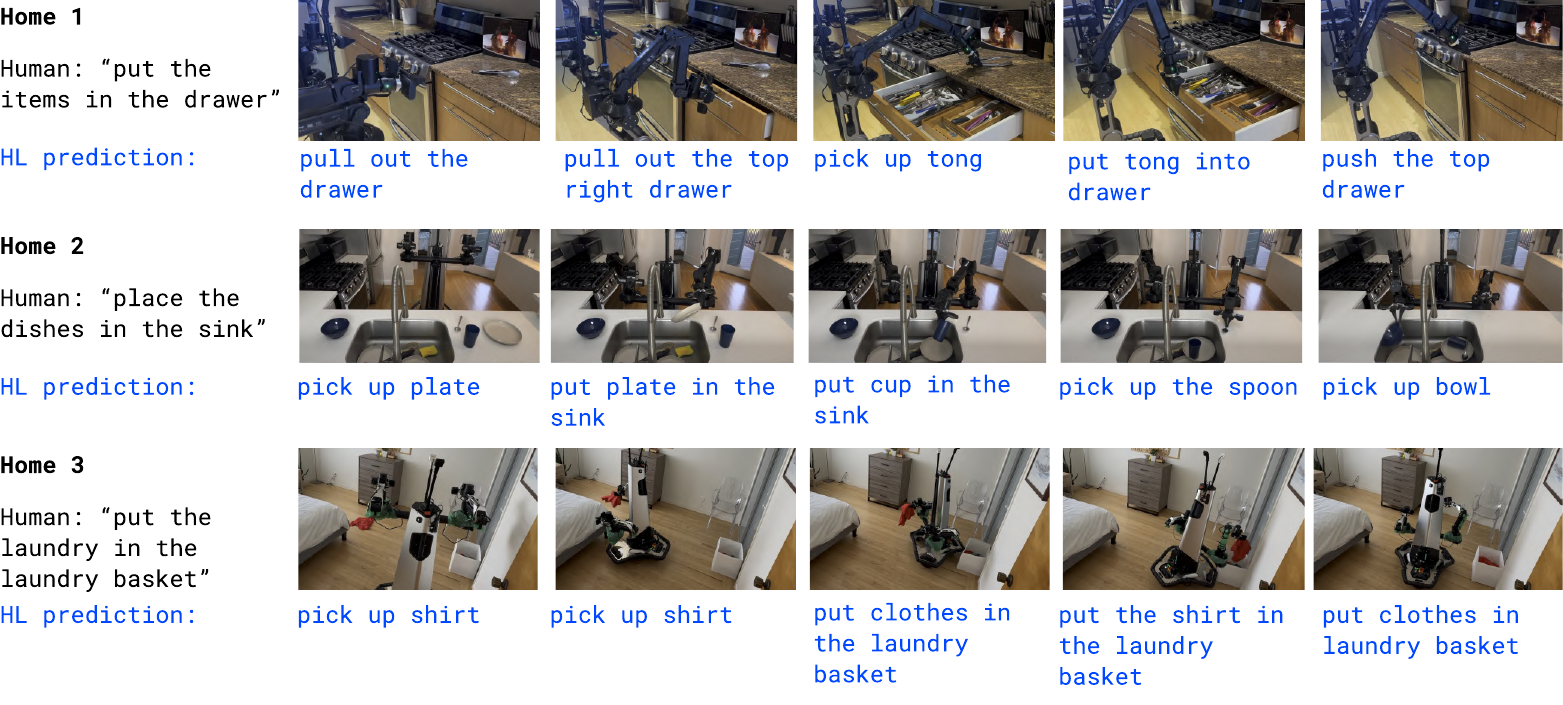}
        \caption{\textbf{Example rollouts.} We visualize an exemplary \ModelSymbol\ episode for one task from each home. Top to bottom: putting items in a drawer in Home 1, followed by putting dishes in the sink in Home 2, and putting clothes in the laundry basket in Home 3. The human instruction for each is given on the left, and the high-level subtask prediction from \ModelSymbol\ is shown beneath each frame in blue.}
    \end{subfigure}%
    \hfill
    \begin{subfigure}{0.42\textwidth}
        \centering
        \includegraphics[width=\linewidth]{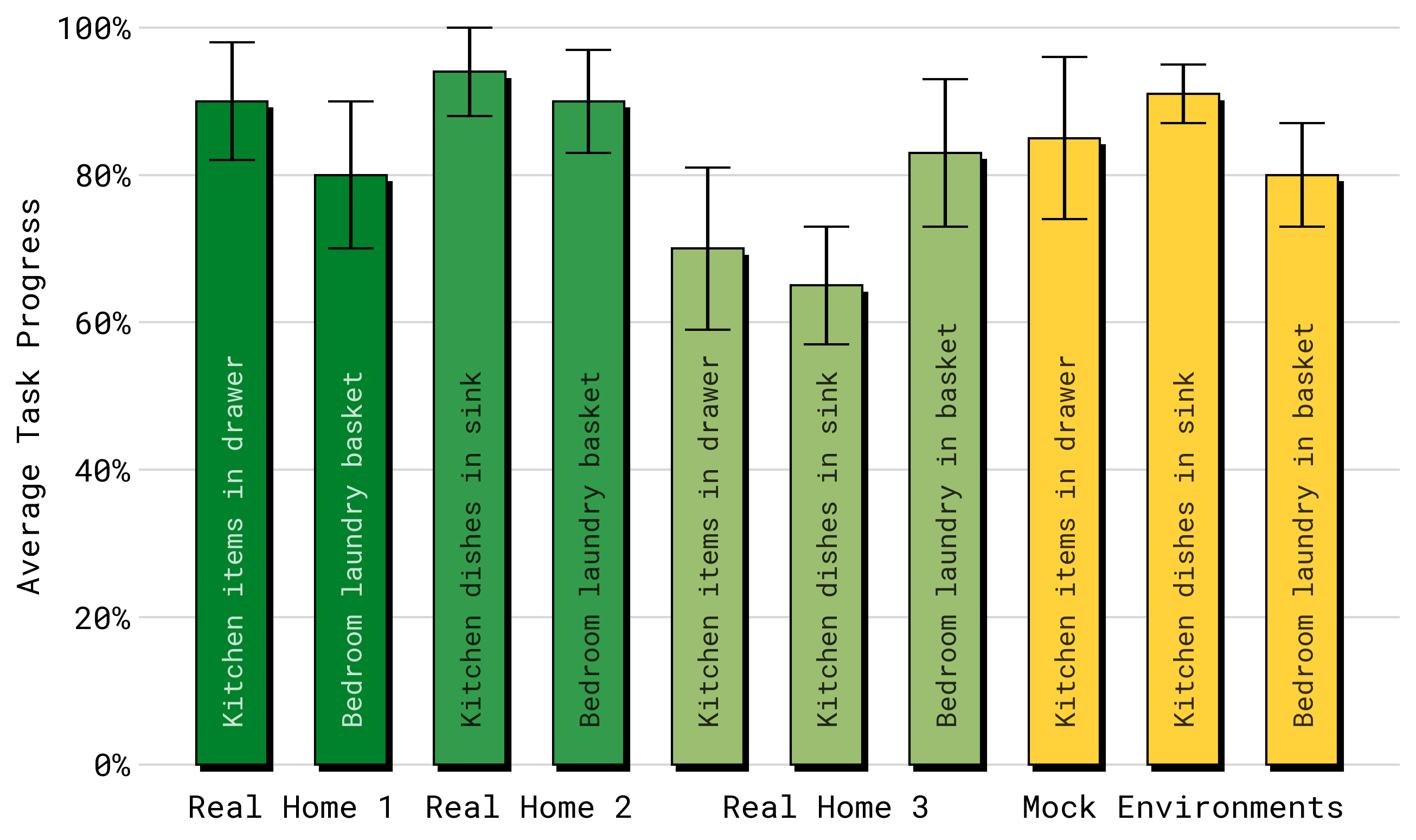}
        \vspace{0.2em}
        \caption{\textbf{Quantitative evaluation.} We show the task progress per task and environment averaged over 10 trials. We find that \ModelSymbol's performance in the mock evaluation setups is representative of its performance in real homes.}
    \end{subfigure}

    \caption{\textbf{Evaluation in real homes.} We evaluated \ModelSymbol\ in three kitchens and three bedrooms in real homes that were not seen during training. We evaluate the tasks `items in drawer', `laundry basket', and `dishes in sink,' and find \ModelSymbol\ to be successful at these tasks in these completely new, real homes.}
    \label{fig:realhomes}
\end{figure*}

To answer Question \textbf{(1)}, we evaluated \ModelSymbol\ in three real homes that were not present in the training set, using both types of robots. In each of the homes, the robots were instructed to perform a bedroom and kitchen cleaning task. The evaluation rubrics for each task are provided in Appendix~\ref{app:scoring_rubric} and roughly correspond to the percentage of steps in each task that were completed successfully (e.g., placing half the dishes in the sink corresponds to around 50\%). The results in Figure~\ref{fig:realhomes} show that \ModelSymbol\ was able to consistently succeed on a variety of tasks in each home (we note that, additionally, the model is capable of performing many more tasks than used in our quantitative evaluation). 
Many of the tasks involve multiple stages (e.g., moving multiple objects) lasting about 2 to 5 minutes. For these trials, the model is provided with a simple high-level command (e.g., ``place the dishes in the sink''), and the high-level inference process autonomously determines appropriate steps (e.g., ``pick up the cup''). This level of in-the-wild generalization goes significantly beyond the results demonstrated with prior vision-language-action models, both in terms of the degree of novelty that the model must handle, and the task duration and complexity.

\subsection{How does generalization scale with the number of scenes?}
\label{sec:envs}
In the next set of experiments, we aim to measure how generalization scales with the number of environments seen in the training data.
We vary the number of environments in the mobile manipulation data and measure its impact on generalization by training with data from 3, 12, 22, 53, 82, and 104 locations.
Since applying the entire pre-training and post-training recipe to each of these datasets is prohibitively compute-intensive, for these experiments we pre-train on the mixture of robot action prediction data \emph{without} mobile manipulation data, and then compare models post-trained on datasets that comprise mobile manipulation data from varying numbers of environments.
While the datasets split by location in principle differ in size, in practice the number of training steps (40k) is chosen such that each model sees the same number of unique data samples, which allows us to control for dataset size when varying the number of locations used within a post-training experiment.

Each model is evaluated in the mock environments shown in Figure~\ref{fig:mock_envs}, which are not seen in training. We conduct two types of evaluations. First, to evaluate overall performance on multi-stage tasks, we use the standard rubric in Appendix~\ref{app:scoring_rubric} and the mock test homes to evaluate each model's end-to-end performance on putting dishes in the sink, packing items into a drawer, putting away laundry, and making a bed. Second, we conduct a more fine-grained evaluation of each model's ability to follow language instructions and interact with novel objects, where the robot must pick up specific objects from a kitchen counter based on language commands. These experiments use both in-distribution objects from similar categories as those in the training data (but new instances), as well as out-of-distribution objects from unseen categories. The latter necessitates broad semantic generalization.

\begin{figure}
    \centering
    \includegraphics[width=\linewidth]{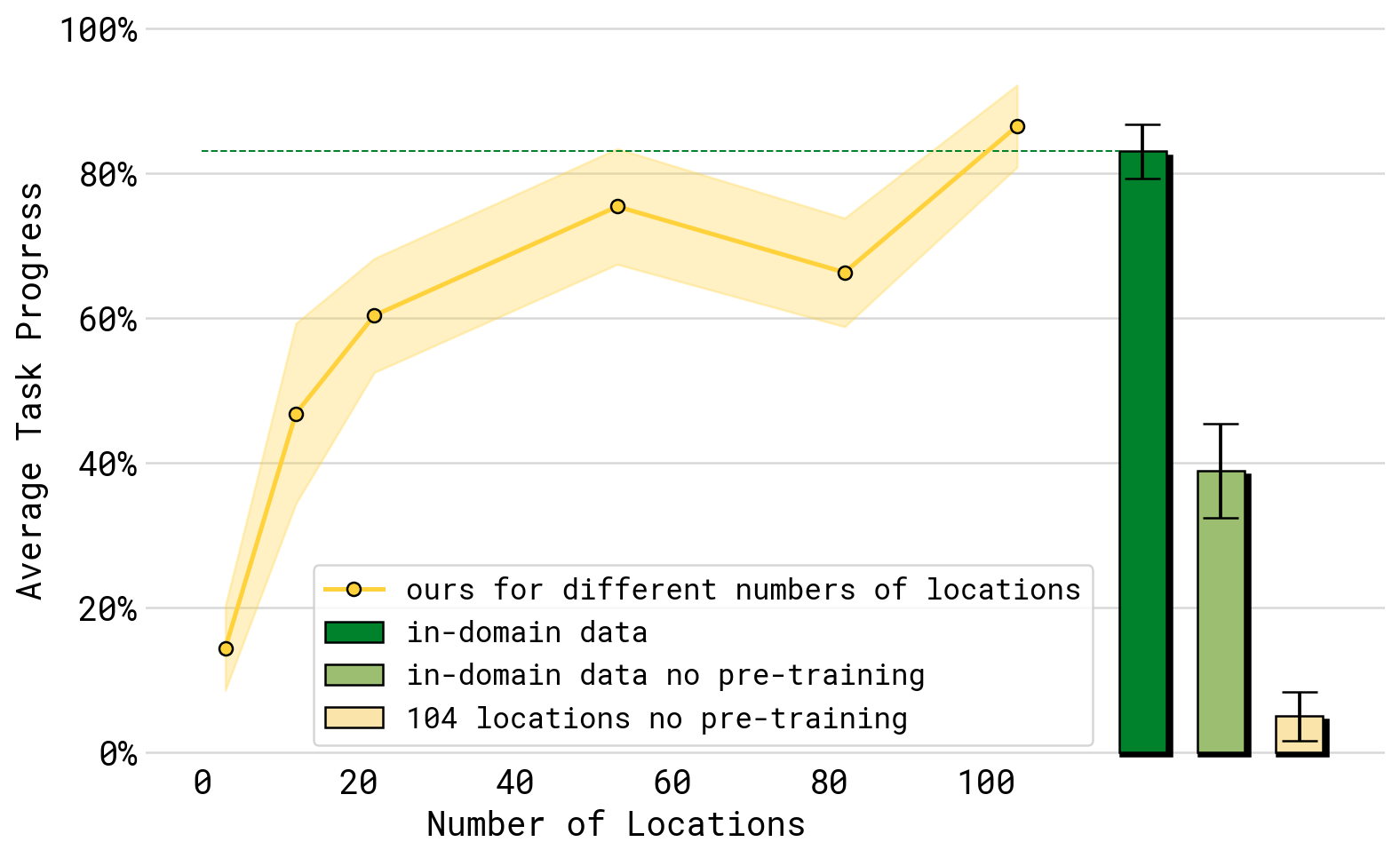}
    \caption{\textbf{Evaluating performance with different numbers of locations.} Performance over the four test tasks --- ``dishes in sink'', ``items in drawer'', ``laundry basket'', ``make bed'' --- improves with more training environments. The dashed green line and green bar show a baseline model that includes the test homes in the training set. Compared to this model, our best model achieves similar performance, despite not seeing any data from the test homes.
    }
    \label{fig:envscaling}
\end{figure}

The results of the first experiment are shown in Figure~\ref{fig:envscaling}. The average performance among the tasks generally improves with more training locations. To quantify how much the final model (with 104 locations) bridges the generalization gap, we include a control (shown in green) that is trained directly on data from the test homes. This control attains similar performance as the final 104-location model, suggesting that our co-training recipe effectively enables broad generalization, reaching similar performance to a model trained on the test environment. To confirm that this generalization performance requires our full co-training recipe, we additionally include two baselines that \emph{do not} use any of the other co-training tasks in the pre-training phase, but instead train directly on either data from the test environment (light green) or mobile manipulation data from the 104 training locations (light yellow). 
The performance for both those baselines is significantly worse --- this indicates that the other data sources leveraged by our full training recipe are essential for good generalization, even when the policy has seen robot data from test homes.
When not using data from test homes, pre-training with our recipe is especially important, as can be seen by the large gap between the green bars and light yellow bar in Figure~\ref{fig:envscaling}.

\begin{figure}
    \centering
    \includegraphics[width=\linewidth]{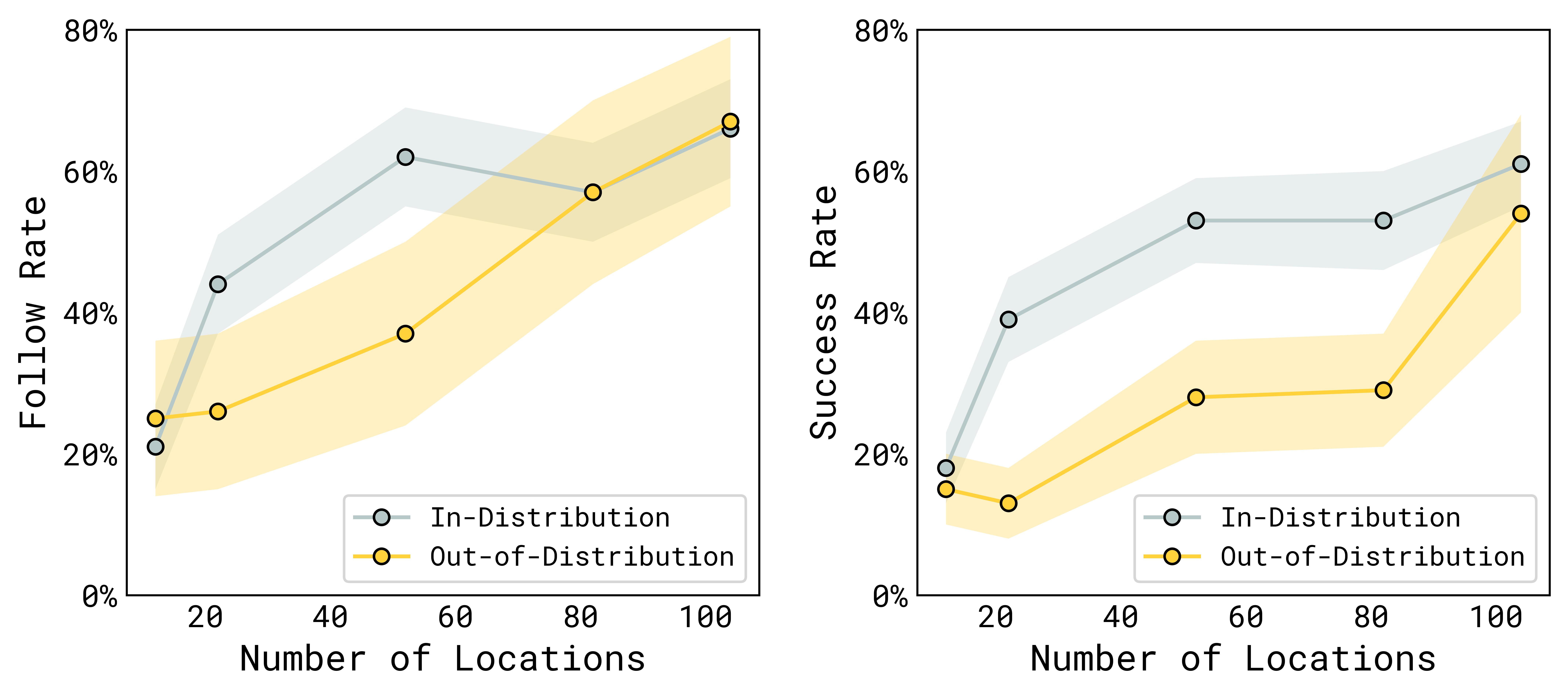}
    \caption{\textbf{Evaluating language following with different numbers of training locations.} We evaluate language following rate and success rate for picking up user-indicated items and placing them into drawers or sinks, averaged over seen object categories (``in-distribution'') or unseen categories (``out-of-distribution''). Performance increases steadily as we increase the number of training locations.
    }
    \label{fig:envscaling_ll}
\end{figure}

The results of the second experiment (language following) are shown in Figure~\ref{fig:envscaling_ll}. We report the language following rate, which measures how often the robot selects the object indicated in the language command, and success rate, which measures how often the robot successfully places that object in the correct location (either inside the drawer or inside the sink, depending on the test scenario). We separately measure performance on object categories seen in training (but new object instances) and unseen (``out-of-distribution'') object categories. 
Details of this experiment are shown and discussed in Appendix~\ref{app:language_following}. Figure~\ref{fig:envscaling_ll} shows that, as the number of locations in the training data increases, both language following performance and success rate improve. As expected, the performance on in-distribution objects improves more quickly than that of out-of-distribution objects. 
As each new environment introduces new household items, the model becomes generally more robust and starts to generalize to task categories that were not present in the training data.

\subsection{How important is each part of our co-training recipe?}

To study Question \textbf{(3)}, we compare our full \ModelSymbol\ model to other training mixtures to study the importance of each mixture component, again using end-to-end task performance in the mock homes and the language following evaluation described in Section~\ref{sec:envs}. As a reminder, our full recipe uses data from mobile manipulators in many environments (\MM), static manipulators in many environments (\ME), and diverse cross-embodiment data collected in laboratory settings (\CE). It also includes high-level data where the prediction corresponds to a high-level language command (\HL), and web data corresponding to captioning, VQA, and object localization tasks (\WD). Post-training also uses verbal instruction data (\VI), which we analyze in Section~\ref{sec:eval_high_level}.
In these experiments, we ablate different parts of the mixture:
\begin{enumerate}
    \item \textbf{no \WD}: this ablation excludes web data.
    \item \textbf{no \ME}: this ablation excludes multi-environment non-mobile data.
    \item \textbf{no \CE}: this ablation excludes the laboratory cross-embodiment data.
    \item \textbf{no \ME\ or \CE}: this ablation excludes both data sources from other robots, such that the model is trained on only data from the target mobile manipulator platform as well as web data.
\end{enumerate}

\begin{figure}
    \centering
    \includegraphics[width=0.9\linewidth]{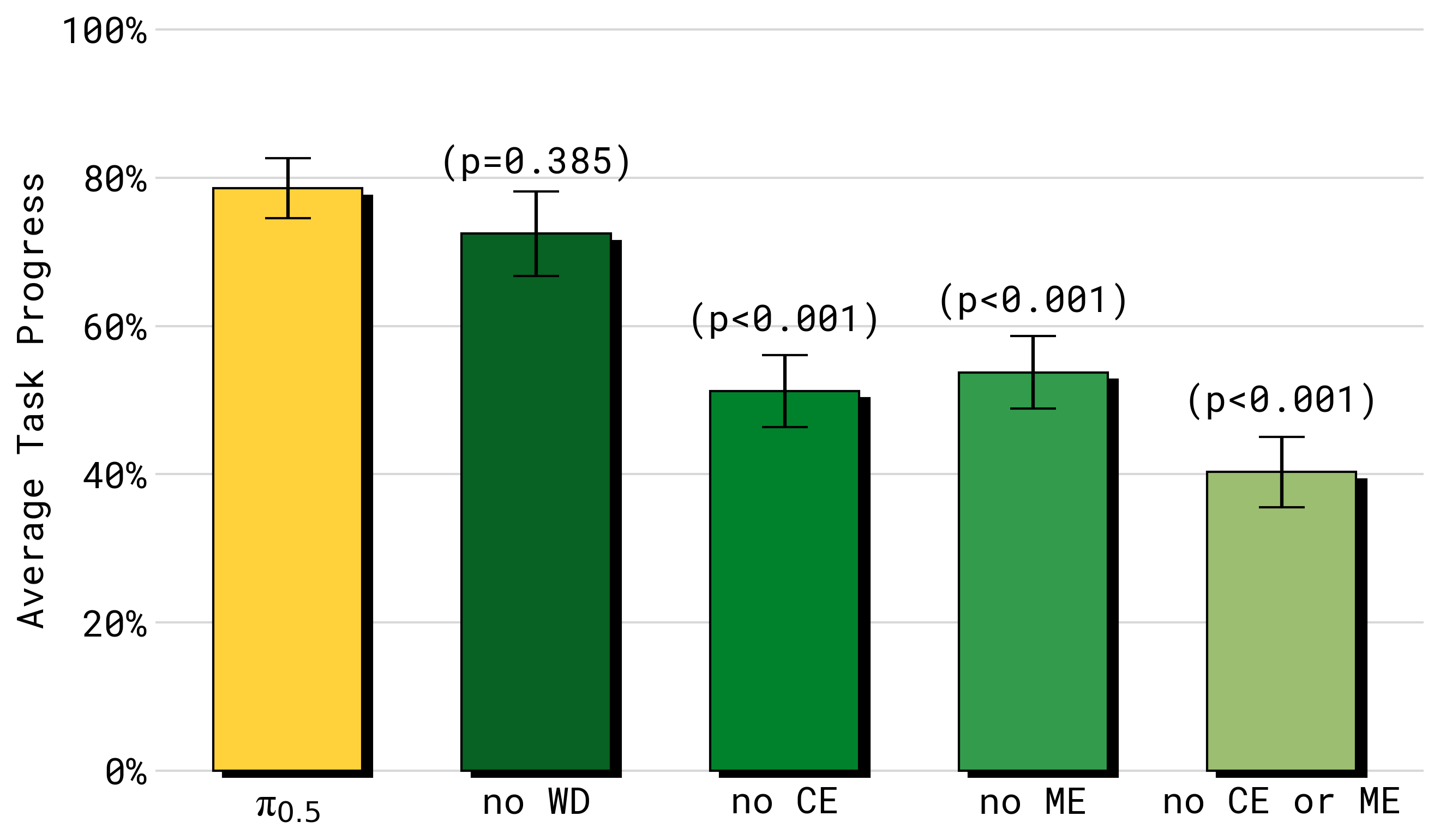}
    \caption{\textbf{Training recipe ablations, mock homes.} We evaluate variants of our model that exclude different parts of the training mixture on all four test tasks (10 trials per policy and task). Including cross-embodiment data, both in diverse environments (\ME) and for diverse tasks in laboratory settings (\CE) is important for good performance, with large degradation when either or both of these data sources are removed. Web data (\WD) does not make a significant difference in these experiments, but we will see in Figures \ref{fig:language_following} and \ref{fig:high_level} that it impacts object generalization and high-level performance.}
    \label{fig:ablations}
\end{figure}

\begin{figure}
    \centering
    \centering
    \includegraphics[width=\linewidth]{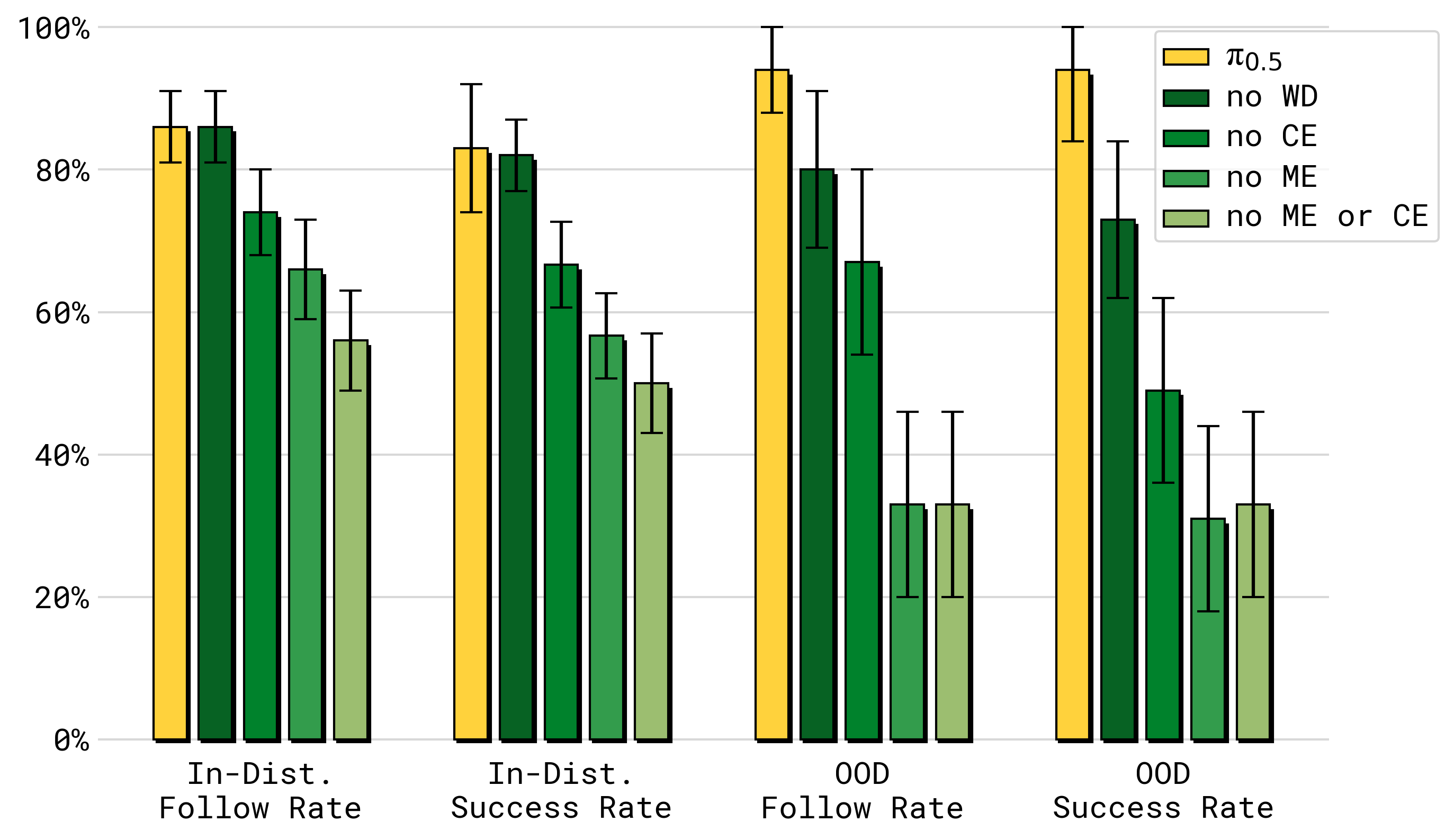}
    \caption{\textbf{Training recipe ablations, language following.} Evaluating language following with in-distribution and out-of-distribution objects after training on different numbers of locations. 
    Including web data (\WD) is important for out-of-distribution (OOD) performance in particular. 
    Cross-embodiment (\CE) and diverse environment (\ME) data both have a large impact on in-distribution and out-of-distribution performance.
    } 
    \label{fig:language_following}
\end{figure}

The results on the full mock home tasks are shown in Figure~\ref{fig:ablations} (detailed breakdown of performance on each task in Appendix~\ref{app:breakdown}). First, we see in the results that excluding \emph{either} of the two cross-embodiment data sources (\ME\ and \CE) significantly degrades performance, indicating that \ModelSymbol\ benefits considerably from cross-embodiment transfer, from both other environments (\ME) and other tasks (\CE). Excluding both sources harms performance even more. Interestingly, the difference in performance with the \textbf{no WD} ablation is not statistically significant in this experiment, though we show later that web data has a large impact on language following (below) and high-level subtask inference (Section~\ref{sec:eval_high_level}).

The results of the language following experiment, 
shown in Figure~\ref{fig:language_following}, show a similar trend as Figure~\ref{fig:ablations} --- excluding \ME\ or/and \CE\ data leads to a significant degradation in performance. What differs now is that removing web data (\textbf{no \WD}) causes significantly worse performance on out-of-distribution (OOD) objects --- we conjecture that training with web data, which contains very broad knowledge of physical objects, allows the model to understand and follow language commands involving \emph{unseen} object categories.

\subsection{How does \ModelSymbol\ compare to other VLAs?}\label{sec:exp_vlas}

\begin{figure}
    \centering
    \includegraphics[width=1.0\linewidth]{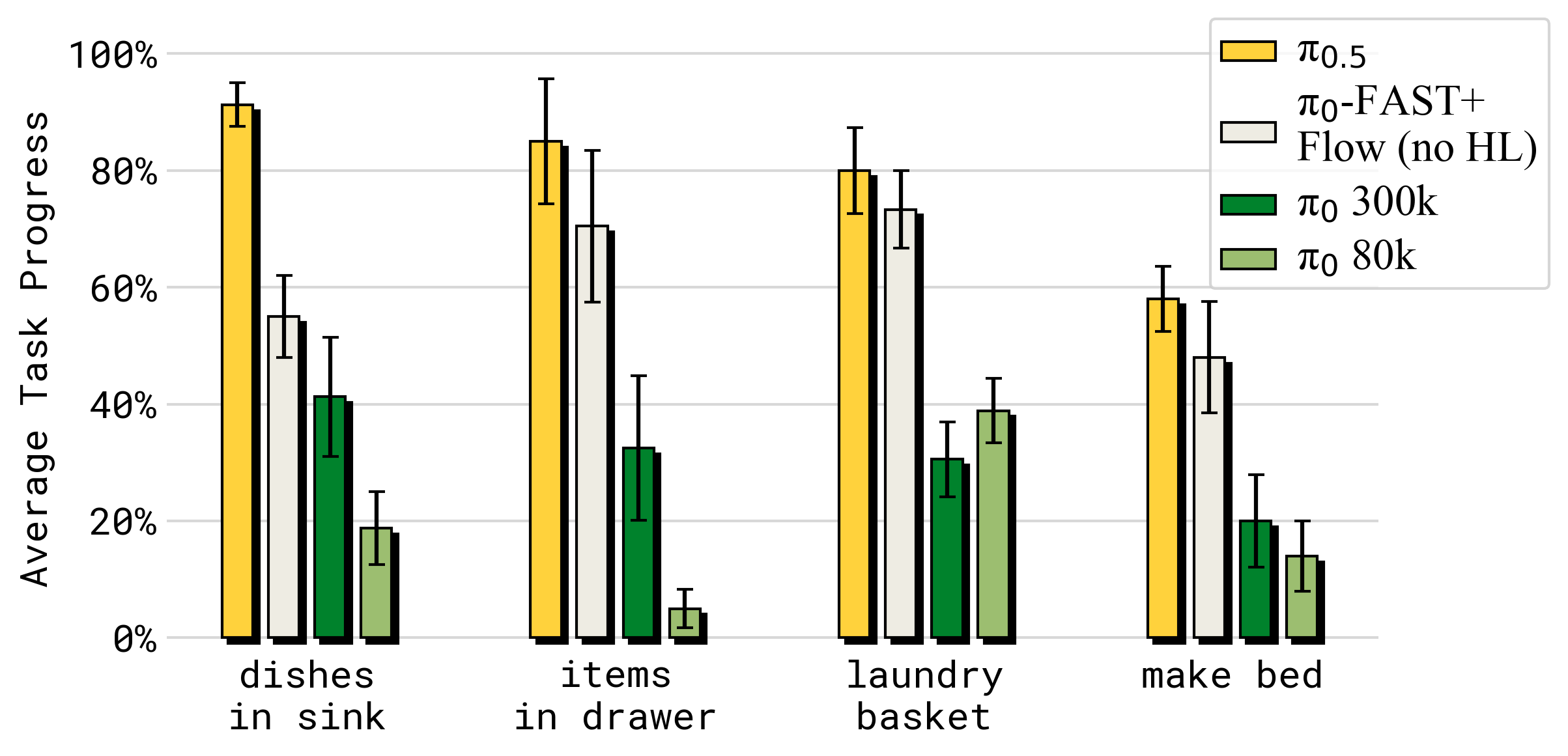}
    \caption{\textbf{Comparing \ModelSymbolBold\ with other models.} Our full model significantly outperforms both \Piz\ and \Pizfast+Flow in the mock home test environments.
    }
    \label{fig:pi0comparison}
\end{figure}

We compare \ModelSymbol\ to the original \Piz\ VLA as well as an improved version of \Piz\ which we denote as \Piz-FAST+Flow. This version is trained via the joint diffusion and FAST action prediction formulation from Equation~\eqref{eq:cotrain}, but on action data only, without the \HL\ or \WD\ datasets. These models provide a strong point of comparison, since \Piz\ has been demonstrated to perform strongly on complex and dexterous mobile manipulation tasks, and the enhancement in \Piz-FAST+Flow brings it as close to \ModelSymbol\ as possible. \ModelSymbol\ builds on these models with a combination of co-training tasks. For a fair comparison, all models receive the same cross-embodiment robot training set and are trained for a comparable number of steps. The differences then are: (1) \ModelSymbol\ additionally uses \HL\ and \WD\ data; (2) \ModelSymbol\ uses a hybrid training procedure, with discrete tokenized training in the pre-training phase, and training with a flow matching action expert \emph{only} in the post-training phase, while \Piz\ always uses the action expert. \Piz-FAST+Flow follows the hybrid training recipe but is trained only with data containing robot actions and thus cannot perform high-level inference. The results in Figure~\ref{fig:pi0comparison} show that \ModelSymbol\ significantly outperforms both \Piz\ and our enhanced version. This result holds even when we allow for longer training up to 300k training steps of \Piz, confirming that as in \citet{pertsch2025fast} training with FAST tokens is more effective in terms of compute than pure diffusion based training. 

\subsection{How important is high-level inference?}
\label{sec:eval_high_level}

Finally, we evaluate the importance of  high-level inference, and compare the performance of several alternative high-level inference methods. The high-level inference mechanism in \ModelSymbol\ takes in a high-level command (e.g., ``clean the bedroom'') and outputs the subtask to complete (e.g., ``pick up pillow''), which is then used as context for inferring the lower-level actions, analogously to chain of thought inference~\citep{wei2022chain}. While \ModelSymbol\ uses a unified architecture where the \emph{same} model performs both high-level and low-level inference, we can also construct
baseline methods that either forego the high-level inference process and feed the task prompt directly into the low-level system, as is common in standard VLA models~\citep{rt22023arxiv,black2024pi_0}, or use another model for high-level inference to ablate the importance of different dataset components in terms of their impact on the high-level policy. We consider the following methods and ablations, all of which use the full \ModelSymbol\ low-level inference process with different high-level policies:

\begin{enumerate}
    \item \ModelSymbol\ model for  high-level and low-level inference.
    \item \textbf{no \WD}: an ablation of \ModelSymbol\ that excludes web data.
    \item \textbf{no \VI}: an ablation of \ModelSymbol\ that excludes the verbal instruction (\VI) data.
    \item \textbf{implicit HL}: no high-level inference at runtime but includes high-level data in training, which may teach the model about subtasks implicitly.
    \item \textbf{no HL}: no high-level inference, and no high-level data in training at all.
    \item \textbf{GPT-4}: use GPT-4 as the high-level policy, evaluating the importance of training the high-level policy on robot data. To align the model with our domain, 
    we prompt GPT-4 with a description of the task and a list of the most used labels to choose from.
    \item \textbf{human HL}: use an expert human as an ``oracle'' high-level policy, to provide an upper bound on performance.
\end{enumerate}

\begin{figure}
    \centering
    \includegraphics[width=\linewidth]{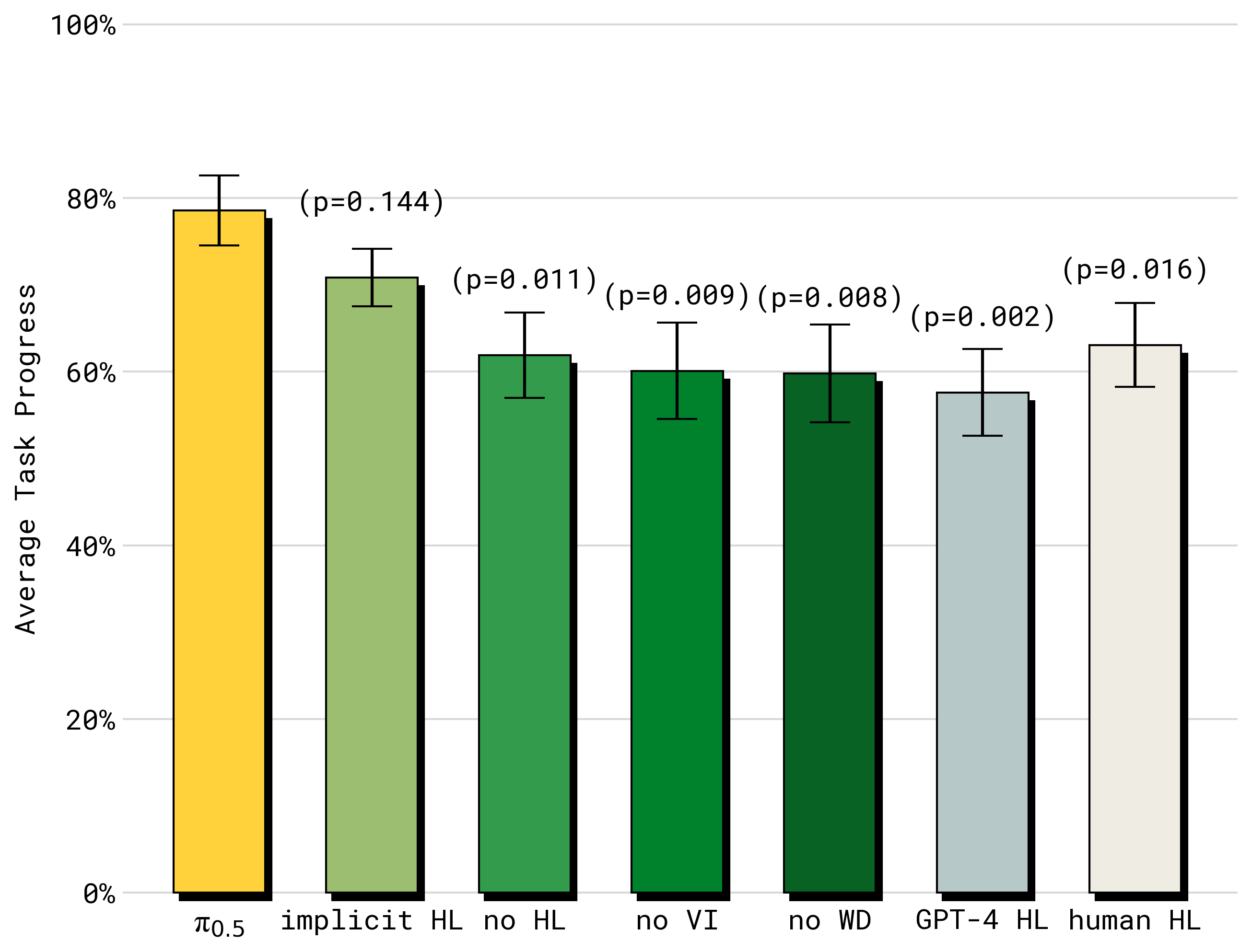}
    \caption{\textbf{Evaluation of the high-level inference process.} While the full \ModelSymbol\ model with high-level and low-level inference attains the best results, using only low-level inference (``implicit HL'') with the full \ModelSymbol\ model also benefits from the inclusion of high-level subtask examples in training. In contrast, excluding verbal instructions (no \VI) or web data (no \WD) leads to a significant degradation in performance, and zero-shot prompting a large API-based model (GPT-4) performs worse.}
    \label{fig:high_level}
\end{figure}

The results of these experiments are shown in Figure~\ref{fig:high_level}. The full \ModelSymbol\ model performs the best, and outperforms even the \textbf{human HL} ``oracle'' baseline. Perhaps surprisingly, the second best model is the \textbf{implicit HL} ablation, which does \emph{not} perform any high-level inference, but includes the full data mixture, i.e.\ also subtask prediction, in training. This strongly suggests the importance of the co-training recipe used by our model: while there is a benefit to explicitly infer high-level subtasks, a significant portion of that benefit is already obtained simply by including subtask \emph{prediction} data in the training mixture. The \textbf{no HL} ablation,  excluding HL task even in training, performs significantly worse. The results also show that the relatively small verbal instruction dataset, which only constitutes about 11\% of the high-level mobile manipulation examples, is critical to strong performance as the \textbf{no \VI} ablation is significantly weaker. The \textbf{no \WD} ablation is also significantly worse, indicating that much of the benefit of web data (perhaps unsurprisingly) lies in improving the high-level policy. Finally, the zero-shot \textbf{GPT-4} ablation attains the worst performance, indicating the importance of adapting VLMs with robot data. 
We provide a detailed breakdown of performance on each task in Appendix~\ref{app:breakdown}, Figure \ref{fig:hl_breakdown}.

\section{Discussion and Future Work}

We described \ModelSymbol, a co-trained model that builds on the \Piz\ VLA to integrate a variety of data sources and enable generalization to new environments. The \ModelSymbol\ VLA can control mobile manipulators to perform tasks in homes that were never seen in the training data, cleaning kitchens and bedrooms, making beds, hanging towels, and performing other multi-stage and dexterous behaviors. \ModelSymbol\ is trained on about 400 hours of mobile manipulation data, but includes a much larger amount of data from other robots, including non-mobile manipulators in diverse environments and data collected under laboratory conditions. It is also co-trained jointly with data from the web, as well as high-level prediction data for outputting language commands based on robot observations. The generalization capabilities of \ModelSymbol\ demonstrate that this co-training recipe facilitates effective transfer, enabling highly generalizable control of a mobile manipulator with only a medium-sized mobile manipulation dataset.

\ModelSymbol\ is not without its limitations. While our VLA exhibits broad generalization, it still makes mistakes. Some environments present persistent challenges (e.g., unfamiliar handles on drawers, or cabinets that are physically hard for the robot to open), some behaviors present challenges with partial observability (e.g., the robot arm occluding a spill that should be wiped), and in some cases the high-level sub-task inference is easily distracted (e.g., closing and opening a drawer multiple times while putting away items). Addressing these challenges with better co-training, transfer, and larger datasets is a promising direction for future work. Other future work directions could address the technical constraints of our method. While \ModelSymbol\ can perform a variety of behaviors to clean up kitchens and bedrooms, it processes relatively simple prompts. The complexity of the prompts that the model can accommodate is determined by the training data, and more complex preferences and instructions could be incorporated by producing more intricate and diverse annotations, either with human labelers or synthetically. The model also uses a relatively modest context, and incorporating richer context and memory could make the model significantly more capable in settings with more partial observability, such as tasks that require navigating between different rooms or remembering where objects are stored. More broadly, \ModelSymbol\ explores a particular combination of heterogeneous data sources, but the specific sources of data can be explored even more broadly. For instance, the ability of our system to learn from verbal instructions provides a powerful new supervision modality, and future work could explore this and other ways that people can provide robots with additional contextual knowledge. We hope that our work will serve as a foundation for a new generation of VLAs that exhibit broad generalization to diverse real-world environments.

\section*{Acknowledgements}

We thank our robot operators for data collection, evaluations, logistics, and video recording. See Appendix~\ref{app:contributions} for a full contributions statement.

\bibliographystyle{plainnat}
\bibliography{references}

\appendix

\section{}

\subsection{Contributions}
\label{app:contributions}

\noindent\textbf{Data collection and operations}. Noah Brown, Michael Equi, Chelsea Finn, Lachy Groom, Suraj Nair, Lucy Xiaoyang Shi, Anna Walling.

\noindent\textbf{Annotation and supplemental data}. Danny Driess, Chelsea Finn, Niccolo Fusai, Lachy Groom, Brian Ichter, Karl Pertsch, Allen Z. Ren, Laura Smith, Kyle Stachowicz, Quan Vuong, Anna Walling, Lili Yu.

\noindent\textbf{Policy training and research}. Kevin Black, Danny Driess, Michael Equi, Chelsea Finn, Niccolo Fusai, Dibya Ghosh, Brian Ichter, Liyiming Ke, Sergey Levine, Suraj Nair, Karl Pertsch, Allen Z. Ren, Lucy Xiaoyang Shi, Laura Smith, Jost Tobias Springenberg, Kyle Stachowicz, Quan Vuong, Homer Walke, Lili Yu.

\noindent\textbf{Policy infrastructure}. Kevin Black, Karan Dhabalia, Danny Driess, Manuel Y. Galliker, Dibya Ghosh, Adrian Li-Bell, Quan Vuong, Haohuan Wang, Ury Zhilinsky.

\noindent\textbf{Robot hardware}. Noah Brown, Adnan Esmail, Tim Jones, Devin LeBlanc, Mohith Mothukuri.

\noindent\textbf{Robot infrastructure}. James Darpinian, Adnan Esmail, Manuel Y. Galliker, Karol Hausman, Szymon Jakubczak, James Tanner.

\noindent\textbf{Writing and illustration}. Kevin Black, Danny Driess, Chelsea Finn, Karol Hausman, Brian Ichter, Sergey Levine, Karl Pertsch, Allen Z. Ren, Lucy Xiaoyang Shi, Jost Tobias Springenberg.

\subsection{Task evaluation rubric}
\label{app:scoring_rubric}
For a quantitative evaluation of our method we performed rigorous evaluation of a subset of four tasks that are included in the training dataset (but evaluated in entirely new scenes and configurations). Among these are two kitchen cleanup tasks and two bedroom cleanup tasks. Each task is evaluated with a consistent set of items for each of the policies within a comparison (but items varied between locations) in three different homes and three different mock kitchens and mock bedrooms respectively (a total of 12 different locations). For each evaluation and each policy, unless otherwise stated, we perform 10 evaluations per task; note that each of these evaluation episodes can span multiple minutes and they are thus time intensive. We present results as percent of total points achieved in each evaluation rubric (as outlined below) and present either per task metrics or metrics averaged across all tasks in four different locations, that are consistent for all policies in a comparison, leading to a total of $40$ evaluations per policy for our standard evaluations. Evaluations were carried out by interleaving execution of policies to control for environmental changes. Some evaluations include cancelled episodes due to robot failures, time limitations or other causes, which are removed. In all cases we control the sample size to be close and report statistical significance according to a two-sided t-test assuming variable number of trials within the plots. The language following evaluations follow a different protocol as described in the main text.                  

The evaluation metrics for the kitchen cleanup tasks, which include placing dishes into a sink and storing items in a drawer, are detailed below.

\begin{itemize}
    \item \textit{Dishes in Sink:} The task begins with 4 dishes (e.g., plates, bowls, cutting boards, utensils) placed near a sink. The robot's goal is to place all of them in the sink. 
    \begin{itemize}
        \item[+1] For each item picked up.
        \item[+1] For each item placed in the sink.
    \end{itemize}
    \textit{Maximum score: 8 points.}
    
    \item \textit{Items in Drawer:} The task begins with an item on a countertop. The robot must place the item into a drawer beneath the counter. 
    \begin{itemize}
        \item[+1] Picking up the object.
        \item[+1] Opening the drawer.
        \item[+1] Putting the object into the drawer.
        \item[+1] Closing the drawer (if the object is inside).
    \end{itemize}
    \textit{Maximum score: 4 points.}
\end{itemize}

Next, we outline the evaluation metrics for the bedroom cleanup tasks: putting laundry away and making a bed.

\begin{itemize}
    \item \textit{Laundry in Basket:} The task begins with an article of clothing lying on the ground. The robot's goal is to pick up the laundry and place it in the laundry basket.
    \begin{itemize}
        \item[+1] Navigating to and picking up the clothing.
        \item[+1] Placing the clothing into or on the laundry basket.
        \item[+1] Clothing is fully inside the basket.
    \end{itemize}
    \textit{Maximum score: 3 points.}
    
    \item \textit{Make the Bed:} The bed starts unmade. The robot must tidy the blanket and place two pillows at the head of the bed.
    \begin{itemize}
        \item[+1] Straightening the blanket so it covers the sheets.
        \item[+1] Placing one pillow at the head of the bed.
        \item[+1] Placing the second pillow at the head of the bed.
        \item[+1] Blanket is straightened very neatly.
        \item[+1] Both pillows are placed very neatly.
    \end{itemize}
    \textit{Maximum score: 5 points.}
\end{itemize}

\subsection{Language following experiment setup}
\label{app:language_following}

\begin{figure}[ht]
  \centering
  \begin{subfigure}[b]{0.15\textwidth}
    \includegraphics[width=\textwidth]{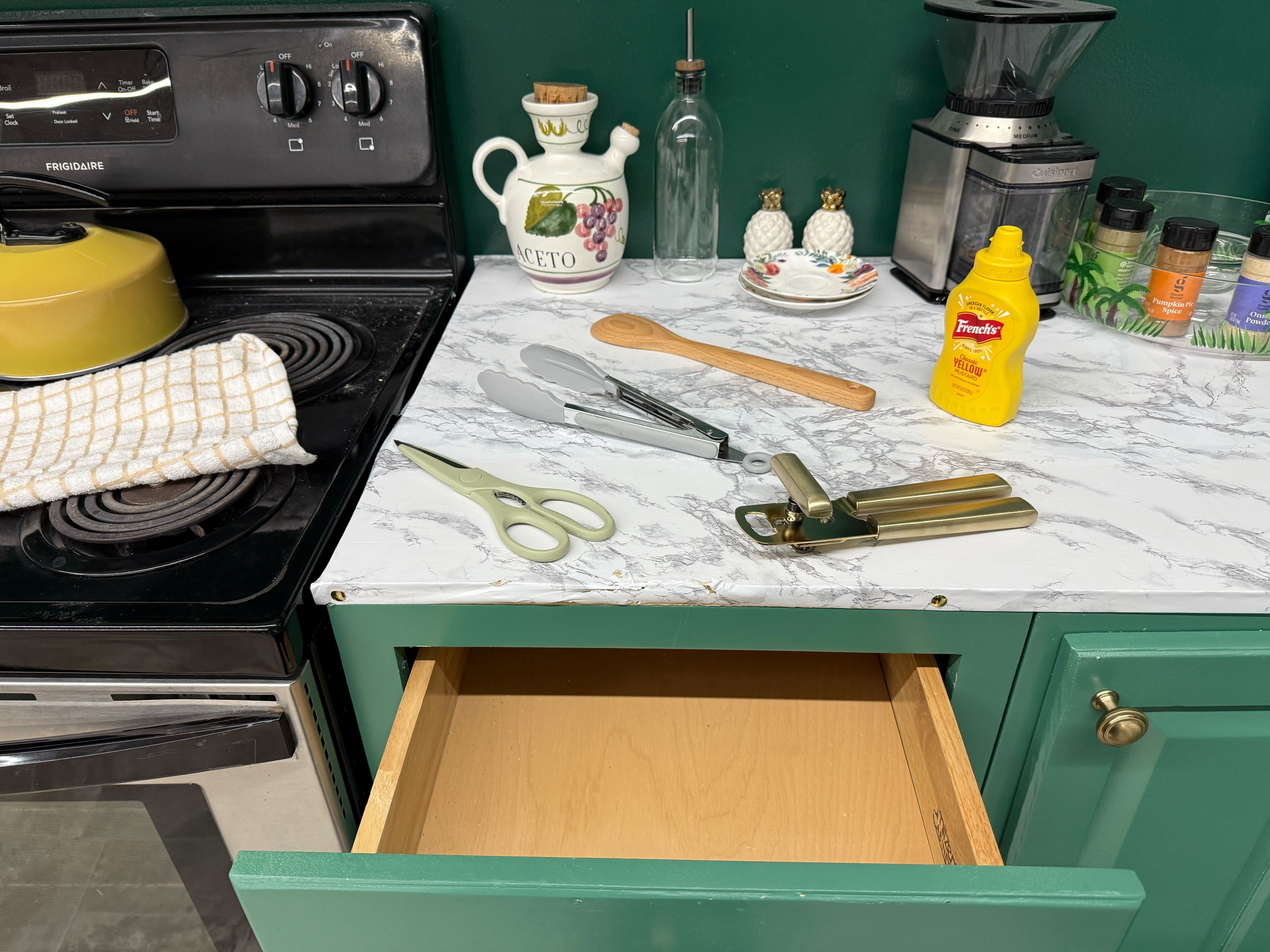}
    \caption{In-distribution objects, items in drawer}
  \end{subfigure}
  \hfill
  \begin{subfigure}[b]{0.15\textwidth}
    \includegraphics[width=\textwidth]{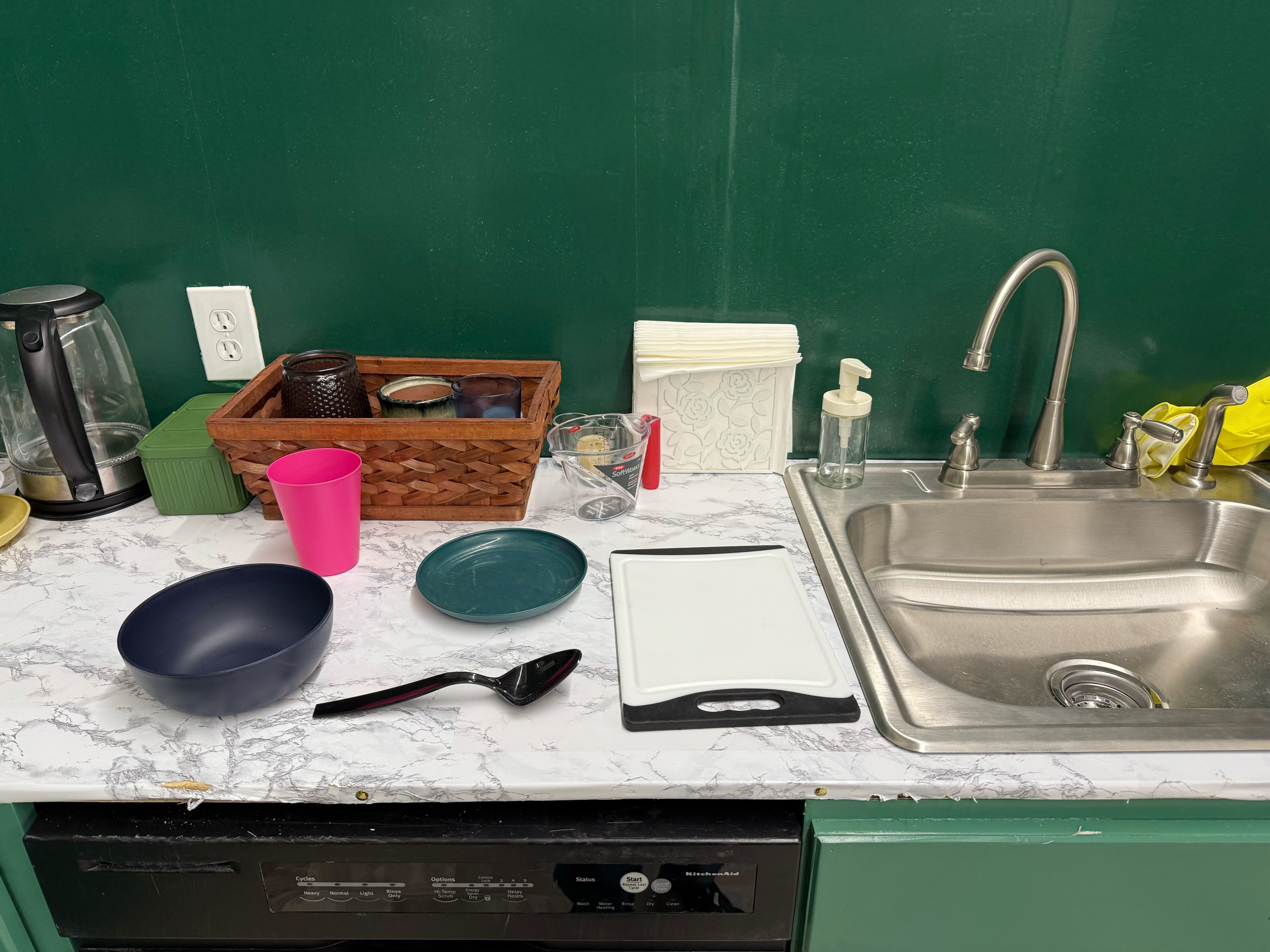}
    \caption{In-distribution objects, dishes in sink}
  \end{subfigure}
  \hfill
  \begin{subfigure}[b]{0.15\textwidth}
    \includegraphics[width=\textwidth]{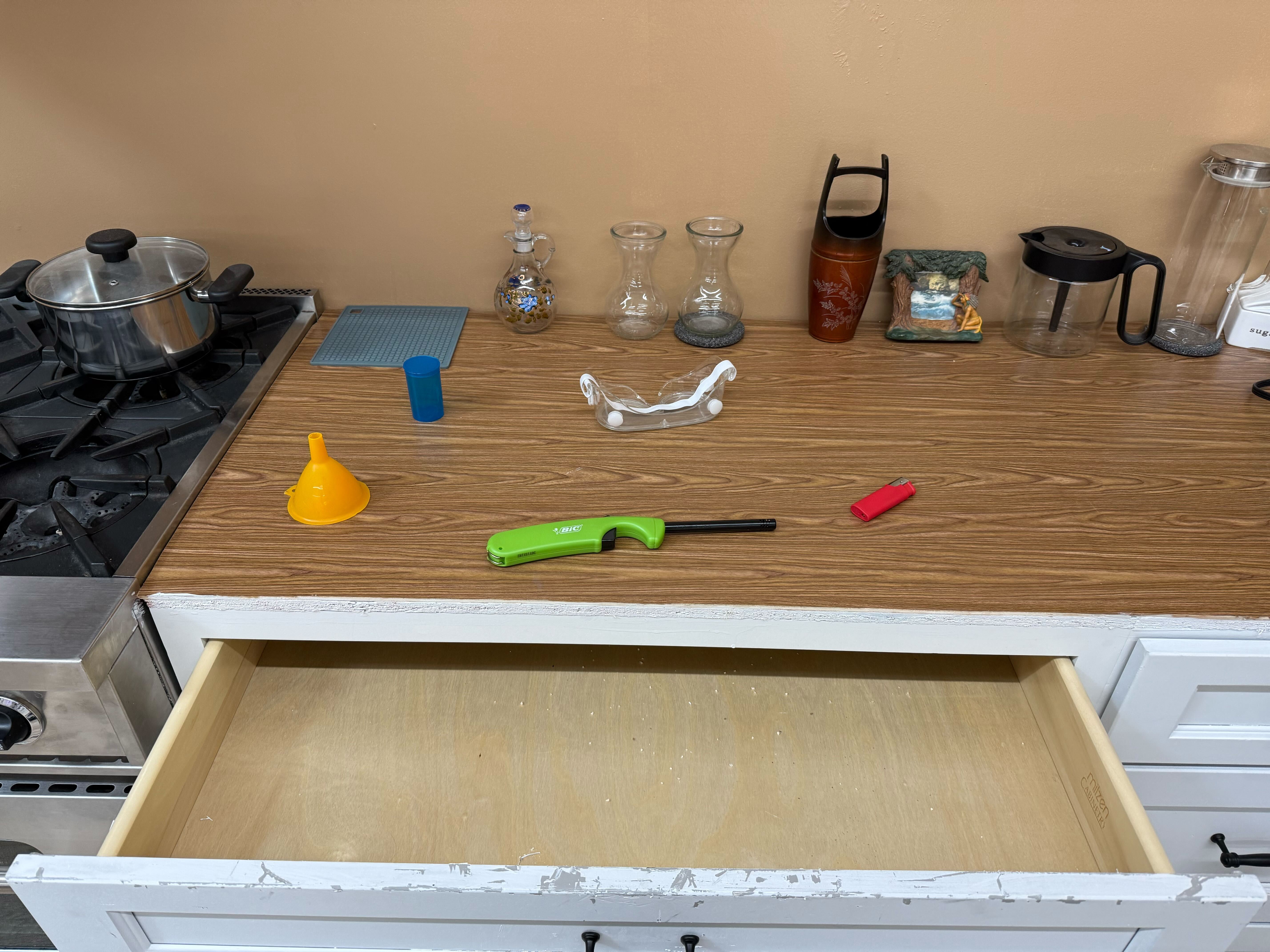}
    \caption{Out-of-distribution objects, items in drawer}
  \end{subfigure}
  \caption{Example initial states of different language following experiments.}
  \label{fig:language_tasks_example}
\end{figure}

The language following experiments use two unseen kitchen scenes to test how well the model follows more specific user commands, such as \textit{``put the scissors in the drawer''} or \textit{``put the cutting board into the sink''}. Each trial requires the robot to interpret the instruction, identify the correct object amidst distractors, and perform the task. We evaluate on two scenarios:
\begin{enumerate}
    \item \textbf{Items in the drawer:} common kitchen items (tongs, wooden serving spoon, can opener, scissors, and small yellow mustard).
    \item \textbf{Items in the sink:} common dining items (cup, bowl, plate, plastic spoon, and cutting board).
\end{enumerate}
In each trial, the robot is presented with five objects and is instructed to move one of them. To discourage shortcut behaviors, the target object is placed further away than the distractors, such that a policy that is unable to interpret the command should achieve only $\sim$20\% language following accuracy. 
We report two metrics, averaged over both scenarios: \textbf{language following rate}, 
which measures whether the correct object was selected, and \textbf{task success rate}, which evaluates whether the object was successfully placed in the specified location.
We further investigate how the number of distinct training environments influences the model’s ability to generalize to previously unseen objects. We design a similar \textbf{Items in the drawer} task with novel household items (a funnel, a pill bottle, a grill lighter, a lighter, and a pair of safety goggles). None of these object categories were present in the training set, ensuring that this task tests the robot’s performance on out-of-distribution objects. We show the example initial scene of each task in Figure~\ref{fig:language_tasks_example}.

\begin{figure}[ht]
    \centering
    \includegraphics[width=\linewidth]{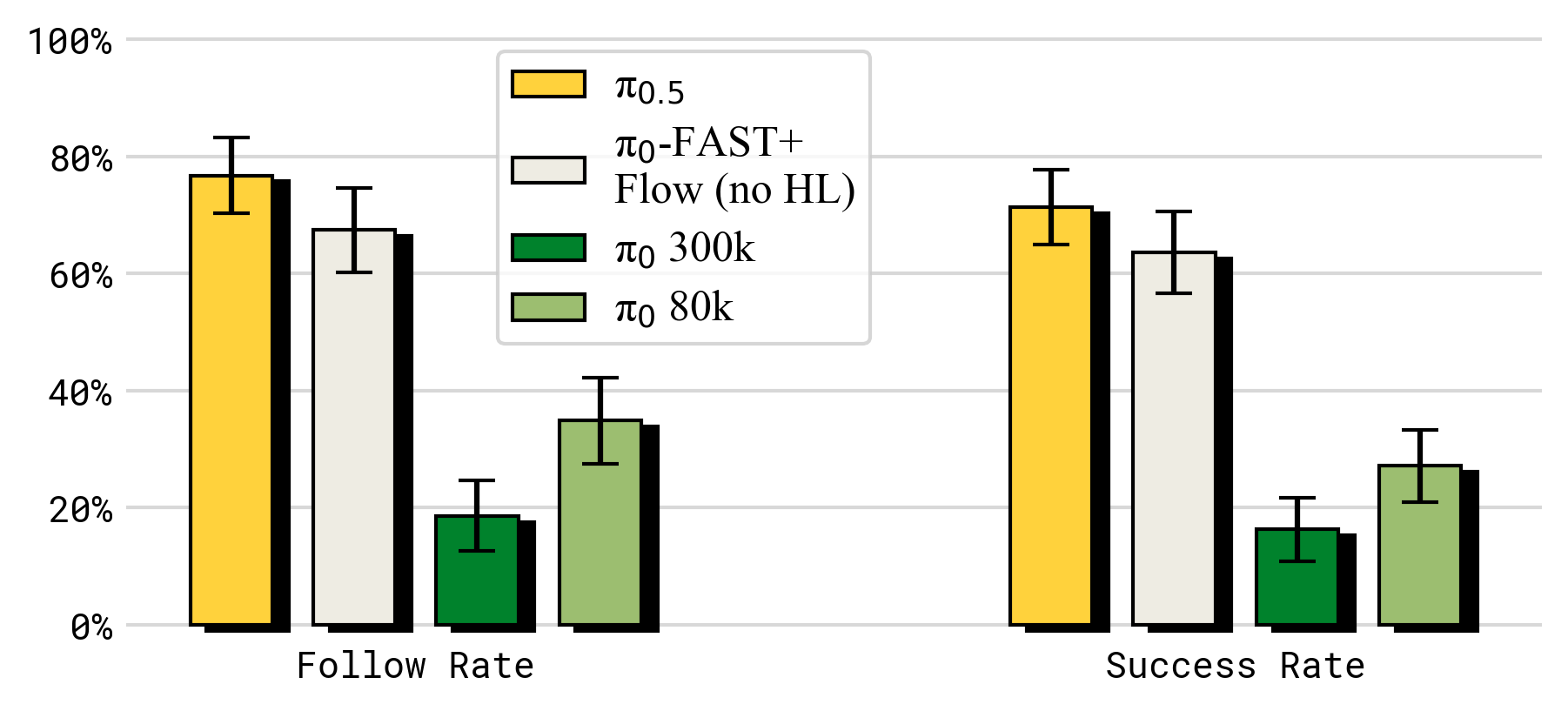}
    \caption{\textbf{Comparing \ModelSymbol\ with other models on language following.} 
    We evaluate language following capabilities of \ModelSymbol\ , \Piz, and \Piz-FAST+Flow, finding \ModelSymbol\ outperforms each, and \Piz\ by a wide margin.
    }
    \label{fig:language_following_model}
\end{figure}

Along with data ablation experiments in Figure~\ref{fig:language_following} and location scaling experiments in Figure~\ref{fig:envscaling_ll}, Figure~\ref{fig:language_following_model} presents language following results across model classes.
We find that \ModelSymbol\ follows language at a slightly higher rate than \Piz-FAST+Flow, and a much higher rate than \Piz\ , indicating the importance of discrete token training on language following abilities.

\subsection{Per-task performance breakdown}
\label{app:breakdown}

\begin{figure}
    \centering
    \includegraphics[width=\linewidth]{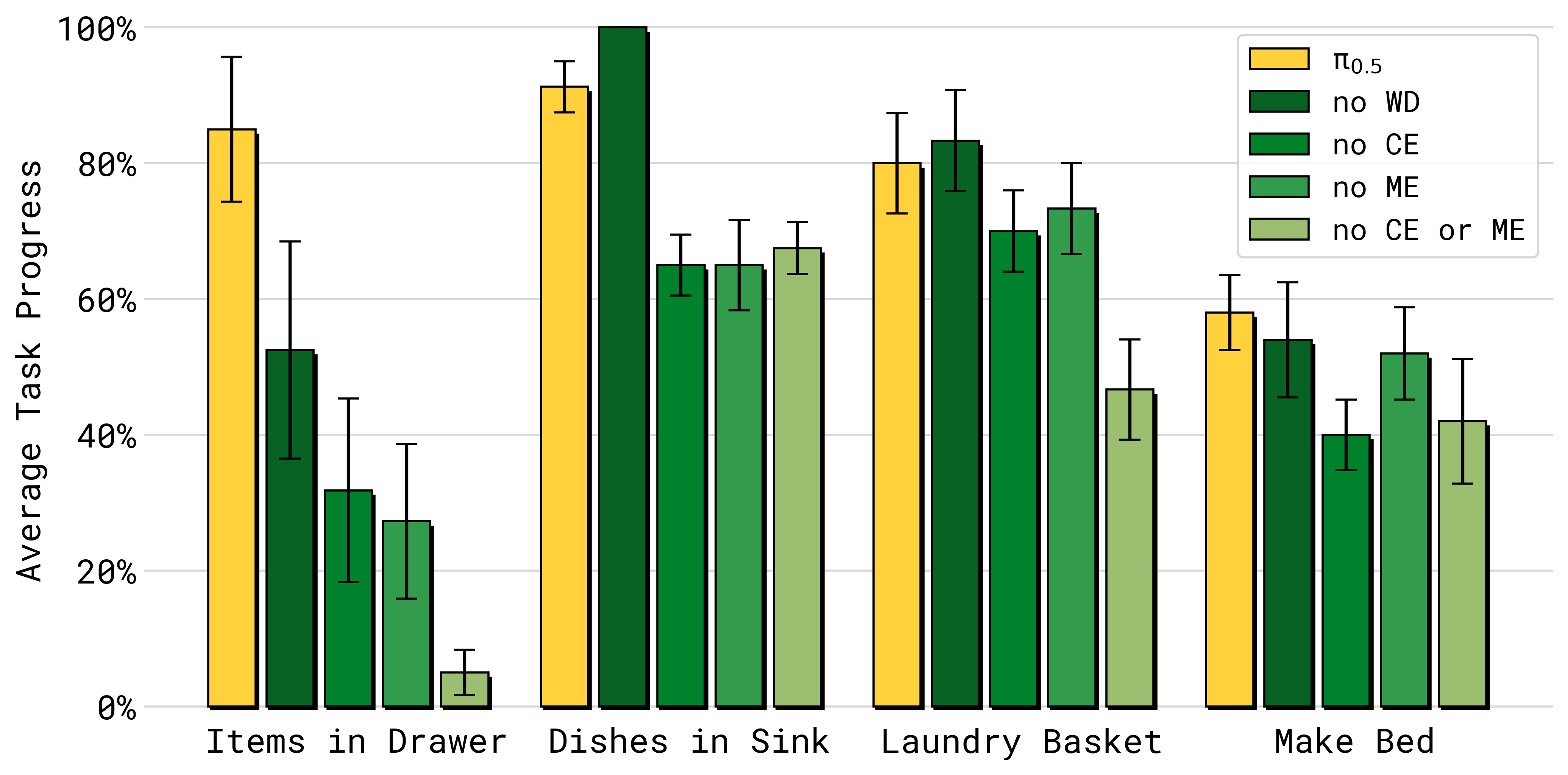}
    \caption{\textbf{Per-task performance breakdown for training recipe ablations.} We evaluate each training mixture variant on four representative household tasks: Items in Drawer, Dishes in Sink, Laundry Basket, and Make Bed. Removing cross-embodiment data (ME or CE) leads to significant degradation in specific tasks, particularly Items in Drawer and Dishes in Sink. Web data (WD) shows greater effect on the task (Items in Drawer) where the broad knowledge of the scene is desired.
    }
    \label{fig:overall_breakdown}
\end{figure}

\paragraph{Co-training recipe ablations}
To better understand the influence of different training data sources on specific task categories, we provide a per-task performance breakdown (Figure~\ref{fig:overall_breakdown}). Here we consider four representative household tasks: \textit{Items in Drawer}, \textit{Dishes in Sink}, \textit{Laundry Basket}, and \textit{Make Bed}. In summary, the results indicate that cross-embodiment transfer and diverse data co-training are critical for generalization across a range of tasks, with varying degrees of reliance depending on task requirements.

For \textit{Items in Drawer}, performance drops substantially when cross-embodiment data (ME or CE) or web data (WD) is removed, with the largest degradation observed when all are excluded. This task requires recognizing and understanding a very broad class of common objects, and such knowledge may be learned from diverse data sources.  
In contrast, \textit{Dishes in Sink} remains relatively robust to the removal of web data (WD) but degrades when cross-embodiment data (ME or CE) is excluded, anchoring the intuition that this task primarily requires general manipulation strategies learned from robotic data.  
\textit{Laundry Basket} and \textit{Make Bed} also exhibit performance degradation when cross-embodiment data is removed, but are generally less sensitive to other changes in the data mixture.

\paragraph{High-level model analysis}
\begin{figure}
    \centering
    \includegraphics[width=\linewidth]{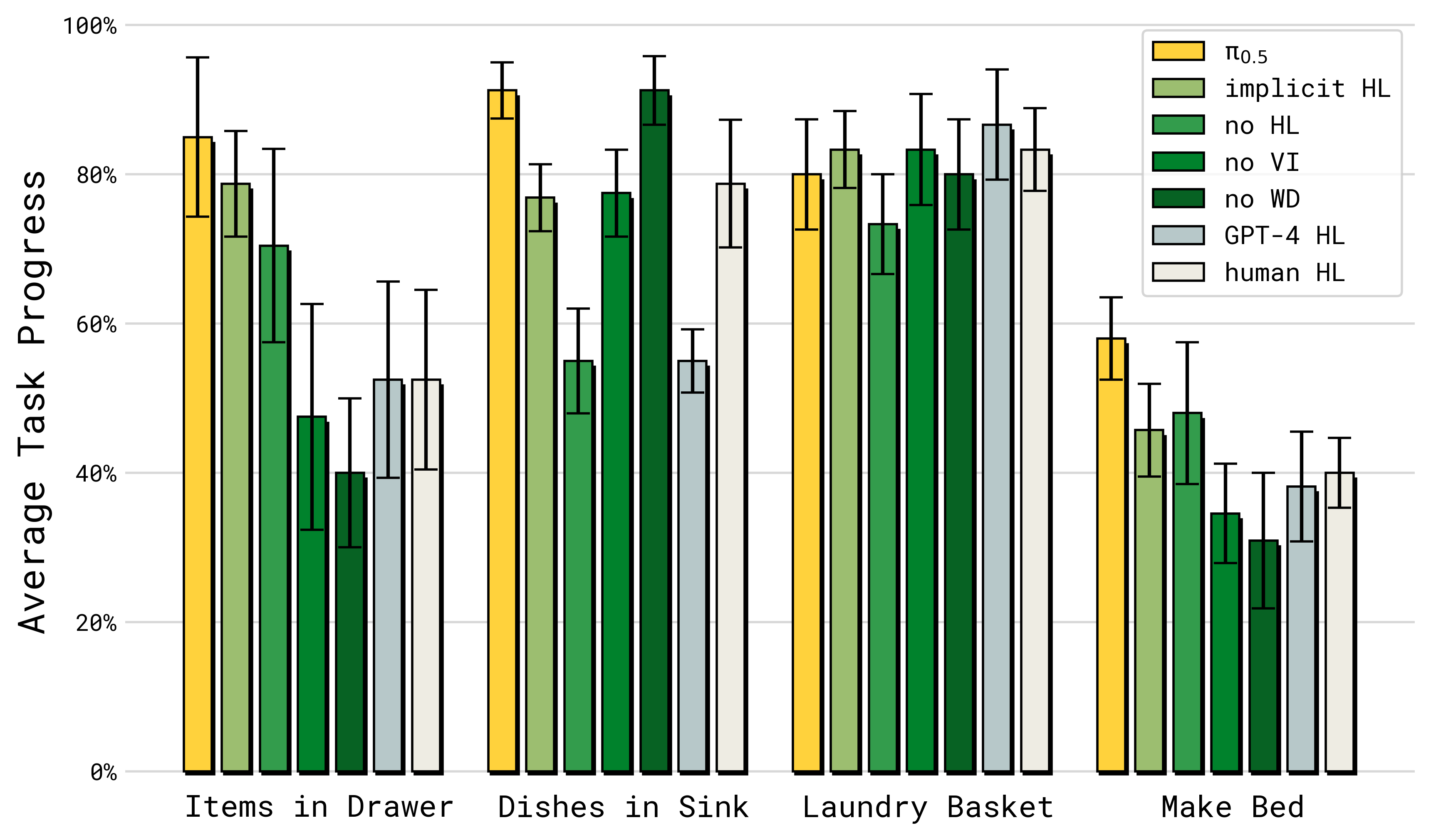}
    \caption{\textbf{Per-task performance breakdown for high-level inference methods.} We evaluate the full $\pi_{0.5}$ model and various high-level inference baselines across four representative household tasks. 
    }
    \label{fig:hl_breakdown}
\end{figure}

For a more granular view of how different high-level inference methods affect specific task categories, we again provide a per-task breakdown (Figure~\ref{fig:hl_breakdown}). We evaluate the full $\pi_{0.5}$ model and all high-level inference baselines across four representative tasks: \textit{Items in Drawer}, \textit{Dishes in Sink}, \textit{Laundry Basket}, and \textit{Make Bed}.
The results show that explicit high-level inference improves performance across tasks, with the full $\pi_{0.5}$ model achieving the best overall results.
  
For \textit{Items in Drawer} and \textit{Dishes in Sink}, high-level inference is critical: performance drops substantially with the \textit{no HL} variant, indicating the importance of structured subtask prediction and long-horizon planning. In these two tasks, the $\pi_{0.5}$ model also outperforms \textit{GPT-4 HL}, showing the benefit of in-domain fine-tuning and demonstrating that the high-level model learns strategies that help the low-level policy succeed.  
In \textit{Items in Drawer}, performance also declines sharply when web data is removed --- this echos the result from the co-training recipe ablation and  highlights the importance of semantic knowledge for generalizing to less seen objects.
For \textit{Laundry Basket} and \textit{Dishes in Sink}, the model is less sensitive to the choice of the high-level policy. These tasks are either relatively shorter in horizon or require less detailed semantic reasoning.

\subsection{Model technical details}
\label{app:model}

The \ModelSymbol\ model builds upon \Piz\ and adopts the PaliGemma VLM~\citep{beyer2024paligemma} as the backbone for visual-language understanding as well as an ``action expert'' for fast action generation.
The VLM backbone takes in a sequence of images $[\mathbf{I}^1_t, \dots, \mathbf{I}^n_t]$ and a language prompt $\lang$ as in \Piz, but also the robot's proprioceptive state $q_t$ in tokenized form and tokenized actions \cite{pertsch2025fast}, which will be auto-regressively predicted.
The action expert is a smaller transformer that takes in a sequence of noisy action tokens $\ba_{t:t+H}^{\tau,\omega}$ for an action horizon of 50, i.e.\ $H=49$, and is trained with the flow matching objective. The noisy action chunk (with action dimension $d$) is first projected to the transformer embedding dimension using a single linear layer. Unlike \Piz\ that fuses the flow-matching timestep $\tau$ with the noisy action before being fed into the transformer, \ModelSymbol\ uses a separate MLP for projecting $\tau$ only and then applies adaptive RMSNorm to inject the timestep information to each layer of the action expert.
The timestep MLP takes in the form of $\mathrm{swish}(W_2 \cdot \mathrm{swish} (W_1 \cdot \phi(\tau)))$, where $\phi : \mathbb{R} \to \mathbb{R}^w$ is a sinusoidal positional encoding function~\citep{vaswani2017attention} and $W_1, W_2 \in \mathbb{R}^{w \times w}$. The action expert outputs action tokens $y^a_{1:H}$, which are then decoded into the target vector field using a final linear projection.

The dimensions of the two transformers are the same as \Piz:
\{\textit{width}=2048, \textit{depth}=18, \textit{mlp\_dim}=16,384, \textit{num\_heads}=18, \textit{num\_kv\_heads}=1, \textit{head\_dim}=256\} for the 2B VLM initialized from PaliGemma weights, and 
the same except for \{\textit{width}=1024, \textit{mlp\_dim}=4096\} for the action expert with 300M parameters.

\begin{figure}
    \centering
    \includegraphics[width=\linewidth]{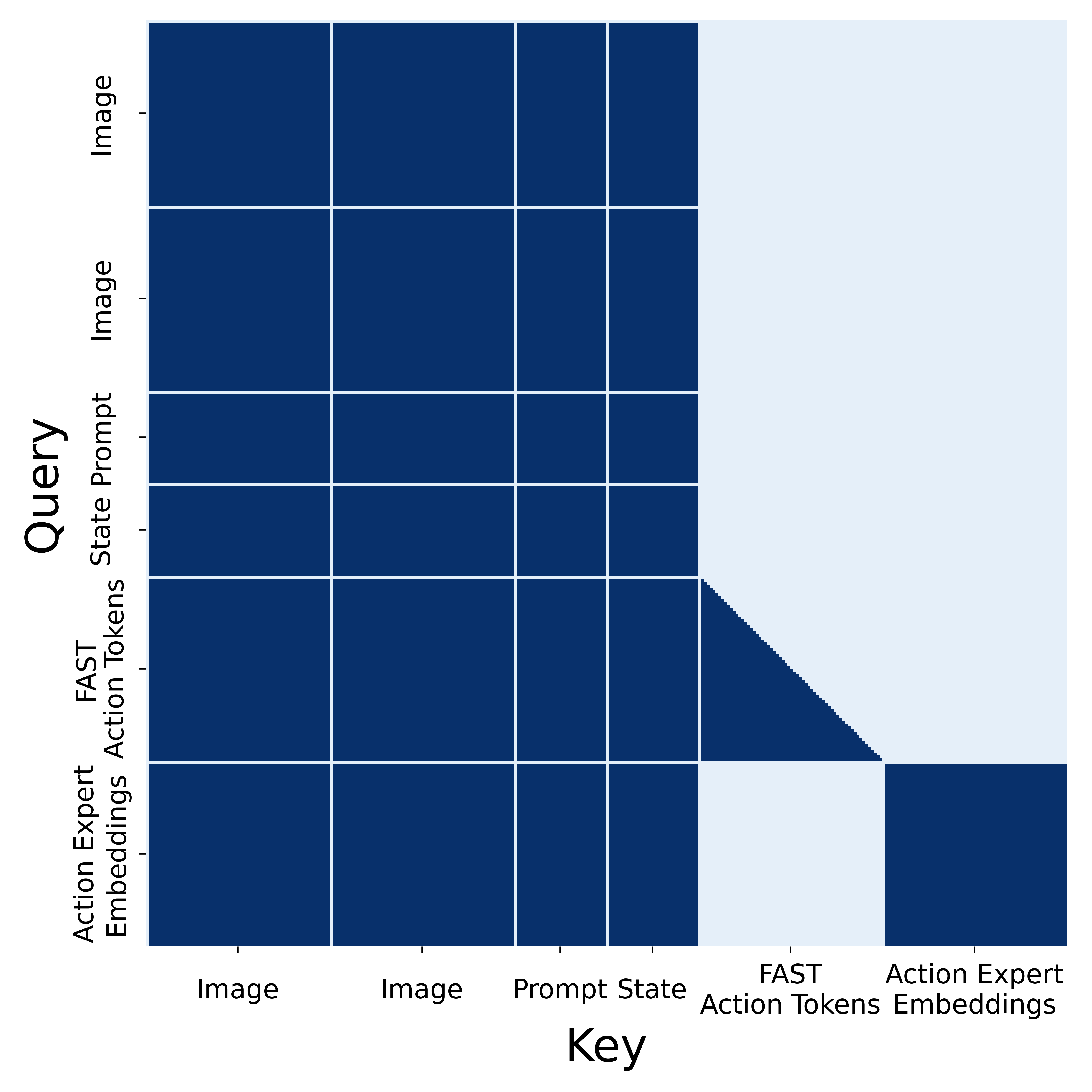}
    \caption{Example of the \ModelSymbol\ attention masking pattern.}
    \label{fig:attention_mask_pattern}
\end{figure}

Embeddings from the VLM and action expert interact only through self-attention. A full prefix mask is used on images, prompt tokens, and proprioceptive state; FAST action tokens attend to this prefix and auto-regressively on previous action tokens. Embeddings from the action expert embeddings attend to the prefix and to one another, but do not attend to FAST action tokens to avoid information leakage between the two representations of actions. In effect, information flows unidirectionally from the VLM to the action expert; no VLM embedding attends to the action expert. An example of the attention mask at each layer is visualized in Figure \ref{fig:attention_mask_pattern}.

We follow \Piz\ for sampling the flow-matching timestep $\tau$. In summary we deviate from standard uniform sampling $\tau \sim \mathcal{U}(0, 1)$~\citep{lipman2022flow,liu2022rectified} or methods emphasizing midrange timesteps~\citep{esser2024scaling}, 
and instead use a time-step sampling distribution that emphasizes low time-steps \citep{black2024pi_0}, given by $p(\tau) = \mathrm{Beta}(\frac{s-\tau}{s}; \alpha=1.5, \beta=1)$. 
Timesteps above the threshold $s$ are excluded from sampling, as they are not needed if the integration step $\delta$ satisfies $\delta > 1-s$. We use $s = 0.999$ in our experiments, which accommodates up to 1,000 integration steps ($\delta > 0.001$).

We apply image augmentation (random crop, resizing, rotation, and color jittering) to all input images using the following hyper-parameters and in this order
\begin{lstlisting}[style=pythonstyle]
transforms = [
augmax.RandomCrop(int(width * 0.95), int(height * 0.95)),
    augmax.Resize(width, height),
    augmax.Rotate((-5, 5)),
    augmax.ColorJitter(brightness=0.3, contrast=0.4, saturation=0.5),
]
\end{lstlisting}

\end{document}